\pdfoutput=1

\documentclass[11pt]{article}

\usepackage[]{acl}

\usepackage{times}
\usepackage{latexsym}

\usepackage[T1]{fontenc}

\usepackage[utf8]{inputenc}

\usepackage{microtype}
\usepackage{inconsolata}
\usepackage{algpseudocode}
\usepackage{algorithm}

\usepackage{xspace} 

\usepackage{subfigure}
\usepackage{tabularx}
\usepackage{multirow}
\usepackage{graphicx}
\usepackage{booktabs}
\usepackage{enumitem}
\usepackage{amsmath}
\usepackage{amsfonts}
\usepackage{amssymb}
\usepackage{txfonts}
\usepackage{bigstrut,rotating}
\usepackage{colortbl}
\definecolor{Gray}{gray}{0.85}

\newcommand\Disco{\textsc{DisCo}\xspace}

\author{\fontsize{12pt}
{\baselineskip}\selectfont Qianren Mao$^{1}$, 
{\bf Weifeng Jiang}$^{2}${\bf, }
{\bf Junnan Liu}$^{4}${\bf, } 
{\bf Chenghua Lin}$^{3}${\bf, } \\
{\bf Qian Li}$^{4}${\bf, } 
{\bf Xianqing Wen}$^{4}${\bf, } 
{\bf Jianxin Li}$^{1, 4 \footnotemark[1]}${\bf, }
{\bf Jinhu Lu}$^{1}${\bf} 
 \\
 \fontsize{10pt}{\baselineskip}\selectfont \textsuperscript{$^{1}$} Zhongguancun Laboratory, Beijing, P.R.China.\\
 \fontsize{10pt}{\baselineskip}\selectfont \textsuperscript{$^{2}$} SCSE, Nanyang Technological University, Singapore.\\
  \fontsize{10pt}{\baselineskip}\selectfont \textsuperscript{$^{3}$} Department of Computer Science, University of Manchester, UK. \\
   \fontsize{10pt}{\baselineskip}\selectfont \textsuperscript{$^{4}$} School of Computer Science and Engineering, Beihang University, Beijing, P.R.China. \\
           \texttt{\fontsize{10pt}{\baselineskip}\selectfont  maoqr@zgclab.edu.cn, weifeng001@e.ntu.edu.sg}
}

\title{Lightweight Contenders: Navigating Semi-Supervised Text Mining through Peer Collaboration and Self Transcendence}

\begin{document}
\maketitle

\begin{abstract}
  The semi-supervised learning (SSL) strategy in lightweight models requires reducing annotated samples and facilitating cost-effective inference. However, the constraint on model parameters, imposed by the scarcity of training labels, limits the SSL performance. 
  In this paper, we introduce PS-NET, a novel framework tailored for semi-supervised text mining with lightweight models. 
  PS-NET incorporates online distillation to train lightweight student models by imitating the Teacher model. It also integrates an ensemble of student peers that collaboratively instruct each other. Additionally, PS-NET implements a constant adversarial perturbation schema to further self-augmentation by progressive generalizing.
  Our PS-NET, equipped with a 2-layer distilled BERT, exhibits notable performance enhancements over SOTA lightweight SSL frameworks of FLiText and \Disco in SSL text classification with extremely rare labelled data.   
\end{abstract}


\section{Introduction}
Deep and sizeable pre-trained language models (PLMs), such as BERT~\cite{DevlinCLT19} and GPT-3~\cite{radford2018improving,BrownMRSKDNSSAA20} have exhibited impressive empirical performance in diverse natural language processing tasks. 
However, the considerable size of these PLMs poses challenges in terms of fine-tuning and online deployment due to latency and cost constraints. Additionally, the efficacy of PLMs in downstream tasks hinges on a fully supervised setup, necessitating abundant manually annotated datasets. Acquiring well-annotated labels is both costly and typically demands domain-knowledgeable professionals. The labour-intensive process of labelling each sentence is susceptible to errors arising from subjective human judgments.

Recent endeavours have been directed towards concurrently addressing two challenges in low-resource applications: \textit{\textbf{how to effectively utilize compressed small models with limited labelled data to achieve model generalization}}. To overcome the challenges of model size reduction and label data scarcity for PLMs, knowledge distillation is employed to compress an original large model (teacher) into a smaller counterpart. Subsequently, lightweight models can be optimized under semi-supervised learning (SSL) conditions using limited labelled data and an abundance of unlabeled data. 
However, two primary  challenges emerge when employing the compressed small model for semi-supervised learning: 
\(\clubsuit\) the scarcity of labelled data samples provides insufficient supervision for the lightweight student model, impeding the acquisition of more nuanced task-specific knowledge, \(\clubsuit\) the smaller compressed model lacks generic regularization for semi-supervised learning, hindering the attainment of enhanced model generalization.

We present PS-NET\footnote{Code and data available at: 
\url{https://github.com/LiteSSLHub/PSNET}.}, a semi-supervised text mining framework that refines lightweight student models using minimal labelled data for effective inference. PS-NET employs supervised optimization with task-specific labelled data, followed by online knowledge distillation to derive student cohorts from unlabeled data.

Within this semi-supervised framework, we incorporate an ensemble of lightweight student models that engage in reciprocal instruction, enabling collaborative optimization through mutual learning.
We also introduce adversarial perturbations to gradually increase learning difficulty, promoting self-improvement in the students. 
In a nutshell, online distillation produces lightweight student networks, while mutual learning and adversarial perturbations refine the optimization process, helping avoid suboptimal solutions and overcome optimization barriers.  These processes enhance the generalization capabilities among lightweight student cohorts.

Extensive experiments across various semi-supervised text classification and semi-supervised extractive summarization tasks substantiate the exceptional performance of lightweight models within our PS-NET framework, even under the constraints of limited labelled data. 
PS-NET showcase fewer parameters (\(12.30\times\) smaller) in comparison to the complete 12-layer BERT models. 
Additionally, our PS-NET, equipped with a 2-layer distilled BERT, surpasses competitors, specifically the state-of-the-art lightweight SSL frameworks FLiText and \Disco, by large margins in performance gains in text classification tasks that utilize only 10 labelled data samples per class.

\section{Background \& Related Work}
\subsection{Semi-Supervised Learning}
Ongoing efforts are dedicated to mitigating the need for extensive human supervision through Semi-Supervised Learning (SSL)~\cite{BoardP89}. The success of SSL methods in the visual domain~\cite{SajjadiJT16,LaineA17,TarvainenV17,QiaoSZWY18,MiyatoMKI19,BerthelotCGPOR19,BerthelotRSCK22,Wang0HHFW0SSRS023,abs-2301-10921} has spurred research interest in the NLP community. Noteworthy techniques, such as UDA~\cite{XieDHL020}, operate under the low-density separation assumption, enabling them to achieve comparable performances to fully supervised counterparts while utilizing only a fraction of labelled samples.  Deep Mutual Learning (DML)~\cite{BoardP89,ZhangXHL18} facilitates knowledge transfer among diverse cohort models, demonstrating superior performance in category recognition domains like image classification tasks~\cite{ZhangXHL18,ParkKYC20,ZhangW0HG022,WangJZL23}.  \citet{ParkKYC20,GuoMSWZZSZ23,HeSCSSM24} show that diversity enhances generalization, and \citet{JiangMLLDYW23} confirm peer-teaching yields superior performance by collaborative learning among cohort models.

\subsection{Faster and Lighter SSL}
In recent years, heightened attention has been directed towards faster and lighter semi-supervised learning (SSL). FLiText~\cite{LiuZFHL21} introduces an inspector network integrated with a consistency regularization framework.  \Disco~\cite{JiangMLLDYW23} stands out as a notable framework. It utilizes a novel co-training technique to promote knowledge sharing among students through diverse data and model views. By employing meticulous data augmentation perturbations, such as adversarial attacks~\cite{KurakinGB17a}, token shuffling~\cite{LeeHLM20}, cutoff~\cite{abs200913818}, and dropout~\cite{hinton2012improving}, it achieves enhanced generalization capability. However, \Disco~\cite{JiangMLLDYW23} necessitates the deployment of large-scale offline models as external sources of knowledge.

\section{Methodology}



PS-NET applies supervised optimization with task-specific labelled data, followed by online knowledge distillation to generate student cohorts from unlabeled data in a phased semi-supervised framework. To strengthen the robustness of these lightweight models, PS-NET integrates \textit{\textbf{peer collaboration}} and \textit{\textbf{self-transcendence}} strategies. Student cohorts engage in mutual learning for collaborative optimization, where multiple students are trained together using complementary knowledge distilled from a shared teacher model. Additionally, we introduce adversarial perturbations~\cite{ZhangNH22,ChenZCH24} to progressively increase learning complexity among those lightweight models, fostering self-improvement and enhancing the generalization capabilities of the student cohorts.

\begin{figure}[t]
	\centering
		\includegraphics[width=0.5\textwidth]{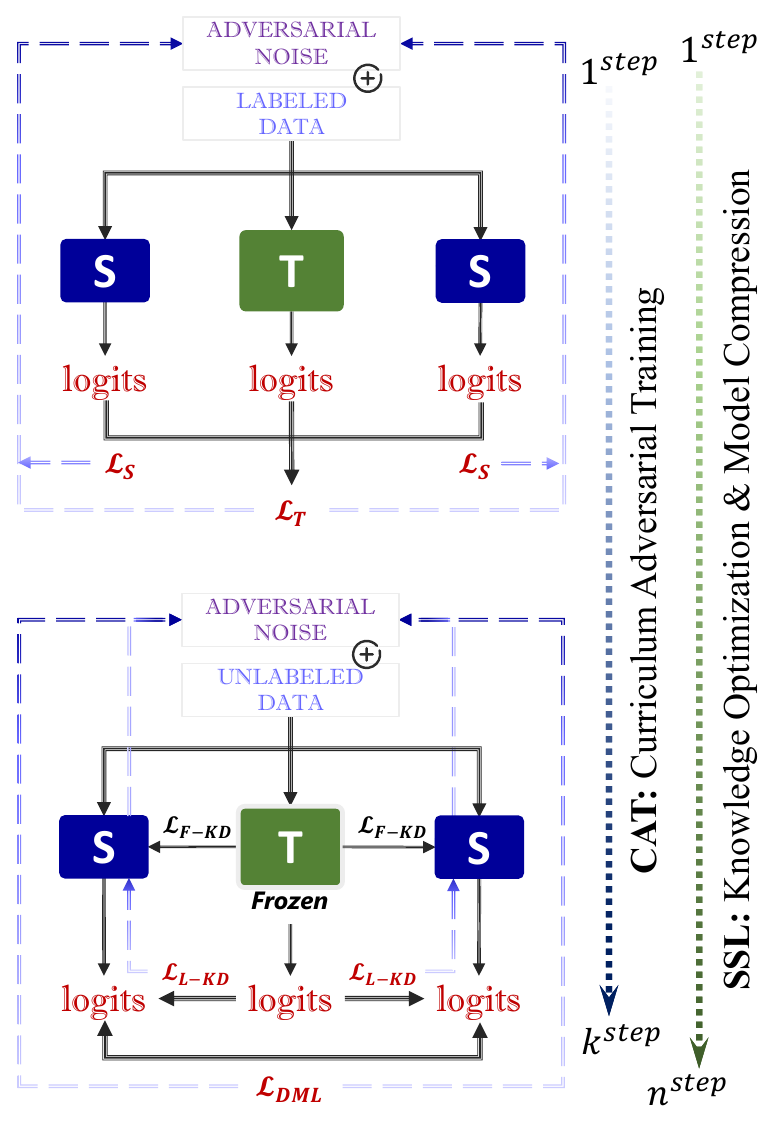}
	\caption{Framework of PS-NET.  It integrates online distillation within an SSL framework, following phased steps of supervised knowledge optimization and unsupervised knowledge distillation. PS-NET allows the student networks to improve generalization through DML in a peer collaboration manner. In each step, PS-NET utilizes CAT, which iteratively generates adversarial noise using gradient-based methods, facilitating continuous self-improvement of the lightweight models.}
	\label{the_framework}
\end{figure}

We illustrate the dual-student PS-NET process for training two lightweight students (see Figure~\ref{the_framework}). Extending this to multiple students is straightforward, as detailed in Section~\ref{multi_students_results}. 
The PS-NET framework is outlined as follows: 
\begin{itemize}[leftmargin=*]
  \item PS-NET involves semi-supervised learning (SSL) of knowledge optimization and knowledge distillation. This sequential approach establishes intermediate objectives that guide the optimization process between student and teacher models.
  \item  PS-NET trains student cohorts through deep mutual learning (DML) by collaboratively mimicking each other's output logits. This enables the cohorts to exchange diversified knowledge, enhancing their generalization ability.
  \item  PS-NET integrates curriculum adversarial training (CAT), which iteratively generates adversarial noise using gradient-based methods. These small perturbations in input embeddings promote continuous self-improvement of student models, reducing their susceptibility to overfitting.
\end{itemize}

\subsection{Knowledge Optimization Procedure}
Formally, given a semi-supervised dataset \(\mathcal{D}\),
\(\mathcal{D}\!=\!\mathcal{S}\cup\mathcal{U}\).
\(\mathcal{S}\!=\!\left \{ (\hat{x},\hat{y})\right \}\) is labeled data and 
\(\mathcal{U}\!=\!\left \{ x^{*}\right \}\) is unlabeled data, and both the student and teacher use all data identically. 
In supervised learning, we employ the cross-entropy loss for optimizing students \({f}^{S}\), and teacher \({f}^{T}\) simultaneously with the labelled data \((\hat{x},\hat{y})\) sampled from \(\mathcal{S}\).

Before being input into the model, the data is augmented by the Curriculum Adversarial Noise Function \(\texttt{ANF}\), as shown in Algorithm \ref{alg:ad_noise}. For labelled data, adversarial noise is generated based on the ground truth labels. The difficulty of the adversarial examples is controlled in a step-wise manner based on the current training step \(n\), gradually increasing the training difficulty from no augmentation to challenging adversarial examples:
\begin{equation}
  \delta_{T}^{\mathcal{S}} = \texttt{ANF}({f}^{T},\hat{x},\hat{y},n),
\end{equation}
\begin{equation}
  \delta_{S}^{\mathcal{S}} = \texttt{ANF}({f}^{S},\hat{x},\hat{y},n),
\end{equation}
\begin{equation}
  \mathcal{L}_{T}=\sum _{(\hat{x},\hat{y}) \in \mathcal{ S}}\texttt{CE}({f}^{T}({\hat{x}+\delta_{T}^{\mathcal{S}}}),\hat{y}),
\end{equation}
\begin{equation}
  \mathcal{L}_{S}=\sum _{(\hat{x},\hat{y}) \in \mathcal{ S}}\texttt{CE}({f}^{S}({\hat{x}+\delta_{S}^{\mathcal{S}}}),\hat{y}).
\end{equation}

\subsection{Model Compression Procedure}
In knowledge distillation, we utilize a uniform function to map the teacher (N layers) and student (M layers, \(M<N\)) through (i) the output of the embedding layer, (ii) the hidden states, and (iii) attention matrices. We set 0 to be the index of the embedding layer. We set \(N+1\) and \(M+1\) to be the index of the prediction layer for the teacher and student.  Hence, the student can acquire knowledge from the teacher by minimizing the MSE objective:
\begin{equation}
  \mathcal{L}_{F-KD}\!=\sum _{x^{*}\in \mathcal{U}}\sum_{m=0}^{M}\texttt{{MSE}}\left ( f_{m}^{S}\left ( x^{*} \right ), f_{g(m)}^{T}\left ( x^{*} \right )  \right ), 
\end{equation} 
where, \(g\left ( m \right )\) is defined as the mapping function between indices from student layers to teacher layers,  indicating that the \(m\)-th layer of the student model emulates information from the 
\(g\left ( m \right )\)-th layer of the teacher model.
The mean squared error loss function (\texttt{MSE}) serves as the distance metric, measuring the similarity of two learned features. We distil the embedding layer, the hidden states, and attention matrices from teacher \(f_{g(m)}^{T}\) to students \(f_{m}^{S}\).

In addition to imitating the behaviours of intermediate layers, we also use knowledge distillation to align the logits of the teacher. The penalty term \(\mathcal{L}_{L-KD}\) is defined as the \texttt{MSE} loss between the logits of the student network's  logits \(f_{M+1}^{S}\) against the teacher's logits \(f_{N+1}^{T}\) over the unlabeled data \(\mathcal{U}\).

Similar to labelled data, unlabeled data also needs to be processed by \(\texttt{ANF}\) before being fed into the model, with difficulty controlled based on the current training step \(n\). However, as shown in Algorithm \ref{alg:ad_noise}, since there is no ground truth label for unlabeled data, we compute the adversarial noise based on the model's prediction of the original data.

\begin{equation}
  \delta_{T}^{\mathcal{U}} = \texttt{ANF}({f}^{T},x^{*},n),
\end{equation}
\begin{equation}
  \delta_{S}^{\mathcal{U}} = \texttt{ANF}({f}^{S},x^{*},n),
\end{equation}
\begin{equation}
\mathcal{L}_{{\tiny L-KD}}\!=\!\!\!\sum _{x^{*}\in \mathcal{U}}\texttt{{MSE}}\left (f_{M+1}^{S}\left ( x^{*} \!+\! \delta_{S}^{\mathcal{U}} \right ), f_{N+1}^{T}\left ( x^{*} \!+\! \delta_{T}^{\mathcal{U}}\right )  \right ).
\end{equation}

After the student cohorts mimic the behaviours of the teacher network from varied perspectives, they then engage in deep mutual learning for collaborative optimization. The student cohorts share diverse knowledge obtained from the teacher and learn from each other, compensating for individual shortcomings to ultimately emulate the teacher. 
Considering two students (\(S\!_{1},S\!_{2}\)) as an example:
\begin{equation}
  \mathcal{L}_{DML}\!\!=\!\sum _{x^{*}\in \mathcal{U}}\texttt{{MSE}}\left (f_{M+1}^{S_{1}}\left ( x^{*} \!+\! \delta_{S_{1}}^{\mathcal{U}}   \right ), f_{M+1}^{S_{2}}\left ( x^{*} \!+\! \delta_{S_{2}}^{\mathcal{U}} \right )  \right ).
\end{equation}

\begin{table*}[tbp]
  \centering
  \footnotesize
  \renewcommand\arraystretch{1}
  \setlength{\tabcolsep}{2mm}{
  \begin{tabular}{@{}lccccrr@{}}
    \toprule
  \textbf{Dataset}       & \textbf{Label Type}               & \multicolumn{1}{c}{\textbf{Classes}} & \multicolumn{1}{c}{\textbf{Labeled}} & \multicolumn{1}{c}{\textbf{Unlabeled}}                                       & \multicolumn{1}{c}{\textbf{Dev}} & \multicolumn{1}{c}{\textbf{Test}} \\ 
  \midrule
  CNN/DailyMail & Extractive Sentences & 2                          & 10/100/1,000                  & \begin{tabular}[c]{c}287,227\\ -10/-100/-1,000\end{tabular} & 13,368                   & 11,490                    \\ 
  \midrule
  AG News       & News Topic               & 4                          
  & \(\times 10/\times 30/\times 200\)                 & 20,000                                                                 & 8,000                     & 7,600                      \\ 
  Yahoo!Answer  & Q\&A Topic                 & 10                         & \(\times 10/\times 30/\times 200\)                 & 50,000                                                                 & 20,000                    & 59,727                     \\ 
  DBpedia       & Wikipedia   Topic        & 14                         & \(\times 10/\times 30/\times 200\)                 & 70,000                                                                 & 28,000                    & 70,000                     \\
  Amzn Review       & Product Review Topic        & 5                         & \(\times 50/\times 200\)                 & 249,000                                                                 & 25,000                    & 65,000                     \\ 
  Yelp Review       & Business Review Topic        & 5                         & \(\times 50/\times 200\)                 & 249,000                                                                 & 25,000                    & 50,000                     \\ 
  \bottomrule
  \end{tabular}
  }
  \caption{Dataset statistics and dataset split for semi-supervised text classification and semi-supervised extractive summarization tasks, in which `\(\times\)' means the number of data per class. `\(-\)' means to subtract the quantity of data.}
  \label{statistics} 
\end{table*}

\noindent\textbf{Overall Training Objective}. Finally, we combine supervised knowledge optimization loss \(\mathcal{L}^{\rm ko}_{\Theta}\) and unsupervised model compression loss \(\mathcal{L}^{\rm mc}_{\Theta}\):
\begin{equation}
\mathcal{L}^{\rm ko}_{\Theta}=\mathcal{L}_{T}+\mathcal{L}_{S},
 \label{loss_dual_6}
\end{equation}
\begin{equation}
  \mathcal{L}^{\rm mc}_{\Theta}\!=\!\mathcal{L}_{KD} + \mu(t,n)\cdot\lambda \cdot \mathcal{L}_{DML},
   \label{loss_dual_7}
  \end{equation}
where \(\mathcal{L}_{KD}=\mathcal{L}_{L-KD} +  \mathcal{L}_{F-KD}\). The term \(\mu(t,n)\!=min(\frac{n}{t},1\)), signifies the ramp-up weight initiating from zero and following a linear curve during the initial \(n\) training steps. The rationale behind incorporating a ramp-up is that the student models, initially initialized randomly, render their mutual learning ineffective.  The hyperparameter \(\lambda\) balances mutual learning. The training of PS-NET involves minimizing the joint loss \(\mathcal{L}^{\rm ko}_{\Theta} + \mathcal{L}^{\rm mc}_{\Theta}\).

\begin{algorithm}[t]
  \caption{Calculation of the Curriculum Adversarial Noise Function \(\texttt{ANF}\)}
  \label{alg:ad_noise}
  
  \textbf{Input:} Model $f$, input embedding $x$, current training step $n$, curriculum step period $\lambda_k$, curriculum step factor $\gamma$, variance of the noise initialization $\sigma^2$, noise boundary $\epsilon$, adversarial gradient ascent learning rate $\eta$, ground truth label $y$ if labeled data is used. \(\lfloor \  \rfloor\) denotes the floor function.
  
  \textbf{Output:} Curriculum adversarial noise $\delta$
  
  \setcounter{ALG@line}{0}
  \begin{algorithmic}[1]
  
  \State $\delta \sim \mathcal{N}(0, \sigma^2)$
  
  \For{$k \gets 0$ to $\lfloor \frac{n}{\lambda_k} * \gamma \rfloor$}
      \If{$y$ exists}
          \State $\delta \gets \delta + \eta \nabla_{\delta} \texttt{CE}(f(x + \delta), y)$
      \Else
          \State $\delta \gets \delta + \eta \nabla_{\delta} \texttt{MSE}(f(x + \delta), f(x))$
      \EndIf
      \State $\delta \gets \Pi_{\|\delta\|_\infty \leq \epsilon} (\delta)$
  
  \EndFor
  
  \State \Return $\delta$
  
  \end{algorithmic}
\end{algorithm}

\subsection{Co-training of Multi-student Peers}
\label{multi_students_results}
PS-NET can seamlessly accommodate multiple students in the cohort. 
Considering \(K\) networks \(\Theta_{1}\),...,\(\Theta_{i}\),...,\(\Theta_{K}\)(\(K\geq 2\)), the objective function for optimising all \(\Theta_{k}\), (\(1\!\leq \!k\leq \!K\)), becomes: 
\begin{equation}
\mathcal{L}^{\rm ko}_{\Theta_{k}}=\mathcal{L}_{T} + \sum _{k=1}^{K} \mathcal{L}_{S}^{k},
 \vspace{-0.15in}
  \label{loss_multi}
\end{equation}
\begin{equation}
  \mathcal{L}^{\rm mc}_{\Theta_{k}}=\sum _{k=1}^{K} \left(\mathcal{L}_{KD}^{k}\!+\!\mu(t,n)\cdot\lambda\cdot\mathcal{L}_{DML}^{k}\right),
  \vspace{-0.05in}
    \label{loss_multi}
  \end{equation}
\begin{equation}
  \mathcal{L}_{DML}^{k}\!=\!\frac{1}{\!K\!-\!1}\!\!\sum _{i=1,i\neq k}^{K}\!\!{\texttt{MSE}} ({f}_{M+1}^{S_{k}}(x^{*}\!+ \delta_{S_{k}}^{\mathcal{U}}),{f}_{M+1}^{S_{i}}(x^{*}\!+ \delta_{S_{i}}^{\mathcal{U}})).
\end{equation}
Equation (\ref{loss_dual_7}) is now a particular case of  (\ref{loss_multi}) with \(k=2\).
When extending the cohort to include more than two networks, a learning strategy for each student of PS-NET takes the ensemble of other \(K-1\) student peers to provide mimicry targets. 
In essence, each student learns individually from all other students within the cohort.

\section{Experiments}
\subsection{Datasets}

As shown in Table~\ref{statistics}, we evaluate PS-NET on semi-supervised extractive summarization and semi-supervised text classification tasks. 

In semi-supervised summarization, models are trained using 100 labelled examples from the CNN/DailyMail dataset~\cite{HermannKGEKSB15}, with the remaining unlabeled examples serving as unsupervised data. We keep the standard splits of the target corpus for validation and testing. 
We conduct experiments on five semi-supervised text classification benchmarks AG News~\cite{ZhangZL15} and Yahoo! Answers~\cite{ChangRRS08} for news classification, and DBpedia~\cite{mendes2012dbpedia} for topic classification, training models with 10, 30, and 200 labelled examples per class. 
We also use the USB benchmark to test Amazon~\cite{McAuleyL13} and Yelp\footnote{\url{https://www.yelp.com/dataset}} review for SSL evaluation, training models with 50 and 200 labelled examples per class.


\subsection{Hyperparameters}
\label{sec:Hyperparameters}
We commence by segmenting sentences through CoreNLP\footnote{\url{https://github.com/topics/corenlp}} and preprocessing the dataset. The source text's maximum sentence length is set to 512 for extractive summarization and 256 for text classification. In the summarization task, we opt for the top 3 sentences from CNN/DailyMail based on the average length of the ORACLE human-written summaries. Fine-tuning on two tasks utilize the Adam optimizer with \(\beta_{1}=0.9\), \(\beta_{2}=0.999\). The supervised learning rates are configured within the intervals [1e-4, 3e-4] and [2e-5, 5e-5] for students and teachers, respectively. The distillation learning and mutual learning rates both range from [5e-4, 7e-4], with the balancing hyperparameter \(\lambda\) set to 0.1. The total number of training steps is 50,000. The warm-up period \(t\) for unsupervised learning is 5,000 steps. For the Curriculum Adversarial Noise Function \(\texttt{ANF}\), the curriculum step period \(\lambda_k\) is 10,000, the curriculum step factor \(\gamma\) is 1, the variance \(\sigma^2\) for initialization is 1e-5, the noise boundary \(\epsilon\) is 1e-6, and the adversarial gradient ascent learning rate \(\eta\) is 1e-3. The supervised batch size is configured as 4, and the unsupervised batch size is set to 16 for classification (32 for summarization) in the majority of our experiments. We optimize hyperparameters through grid search on the development set, selecting the configuration yielding the best validation performance within the initial 10,000 training steps. These optimized hyperparameters are then applied to the complete training process. Model evaluation is performed every 100 training steps for both summarization and classification tasks.


\subsection{Evaluation Methodology}
We evaluate summarization quality using ROUGE 
F1~\cite{LinH03}. We report the full-length F1-based ROUGE-1, ROUGE-2, and ROUGE-L (R-1, R-2 and R-L) on the CNN/DailyMail. These ROUGE scores are computed using \texttt{ROUGE-1.5.5.pl} script\footnote{\url{https://github.com/andersjo/pyrouge}}. 
We report the accuracy (Acc) results for the text classification tasks.

\subsection{Implementation Details}
For the extractive summarization, we can formulate it as a sequence labelling task as ~\citep{LiuL19}. 
The extractive goal is to predict a sequence of labels \(s_{1},...,s_{n}\) \((s_{i}\in{0,1})\) for sentences in a document, where \(s_{i}\!=\!1\) represents the \(i\)-th sentence should be included in the summaries. For the text classification, we feed the hidden state corresponding to each instance's [CLS] token to a softmax classification layer. We used the BERT\footnote{\url{https://github.com/google-research/bert}} for text tokenization. 
For the mapping function \(g(m)\), in the case of two student models, PS-NET students (S\(^{\rm AK}\) and S\(^{\rm BK}\)) are distilled from the first and last \(K\) layers of the teacher model. For scenarios involving multiple student models, additional PS-NET students (S\(^{\rm CK}\), S\(^{\rm DK}\), S\(^{\rm EK}\), and S\(^{\rm FK}\)) are distilled from the intermediate \(K\) layers of the teacher model. 


Each setting runs 3 different random seeds and computes the average performance. The standard deviation of all experimental results falls within the range of [0.3-0.7], which is not displayed to align with the baselines.
All our experiments are conducted on a single NVIDIA Tesla V100 32GB GPU with PyTorch. Notably, our framework generates \textit{two or more} students, from which \textit{one} is selected as the inference model. 
To provide a clearer illustration, we present the performance of each individual student model. 
In practical applications, the selection of a single student model for inference relies on validation set results, ensuring computational efficiency. 
This practice of choosing a single model based on validation performance is standard in methodologies utilizing multiple models, such as mutual learning~\cite{LaineA17,ZhangXHL18,QiaoSZWY18,KeWYRL19}.



\begin{table*}[t]
  \centering
  \footnotesize
  \renewcommand\arraystretch{1.05}
  \setlength{\tabcolsep}{2.7mm}{
    \begin{tabular}{@{}l|l|ccc|ccc|ccc|c@{}}
  \toprule
  \multirow{2}{*}{\textbf{Models}} & \multicolumn{1}{l}{\multirow{2}{*}{\textbf{\(\mathcal{P}\)(M)}}} & \multicolumn{3}{|c|}{\textbf{AG News}}                                          & \multicolumn{3}{c|}{\textbf{Yahoo!Answer}}                                  & \multicolumn{3}{c}{\textbf{DBpedia}}
  & \multicolumn{1}{|c@{}}{\multirow{2}{*}{Avg}}                                   \\ \cmidrule(l){3-11}
  &
  & \multicolumn{1}{c|}{10}    & \multicolumn{1}{c|}{30}    & 200
  & \multicolumn{1}{c|}{10}    & \multicolumn{1}{c|}{30}    & 200   & \multicolumn{1}{c|}{10}    & \multicolumn{1}{c|}{30}    & 200
  &\multicolumn{1}{c}{}  \\
  \midrule
  BERT\(_{\rm BASE}\)
  & 109.48
  & \multicolumn{1}{c|}{81.00}
  & \multicolumn{1}{c|}{84.32}
  & 87.24
  & \multicolumn{1}{c|}{60.10}
  & \multicolumn{1}{c|}{64.13}  & 69.28  & \multicolumn{1}{c|}{96.59}  & \multicolumn{1}{c|}{98.21}  & 98.79
  & 82.18
   \\
  UDA
  & 109.48
  & \multicolumn{1}{c|}{84.70}  & \multicolumn{1}{c|}{86.89}  & 88.56
  & \multicolumn{1}{c|}{64.28}  & \multicolumn{1}{c|}{67.70}  & 69.71  & \multicolumn{1}{c|}{98.13}  & \multicolumn{1}{c|}{98.67} & 98.85 & 84.17 \\
  \midrule
  TinyBERT\(^{6}\)         & 66.96
  & \multicolumn{1}{c|}{71.45}
  & \multicolumn{1}{c|}{82.46}
  & 87.59
  & \multicolumn{1}{r|}{52.84}
  & \multicolumn{1}{c|}{60.59}
  & 68.71
  & \multicolumn{1}{c|}{96.89}
  & \multicolumn{1}{c|}{98.16}
  & 98.65
  & 79.70
  \\
  UDA\(\rm_{TinyBERT^{6}}\)
  & 66.96
  & \multicolumn{1}{c|}{73.90}
  & \multicolumn{1}{c|}{85.16}
  & 87.54
  & \multicolumn{1}{c|}{57.14}
  & \multicolumn{1}{c|}{62.86}
  & 67.93
  & \multicolumn{1}{c|}{97.41}
  & \multicolumn{1}{c|}{97.87}
  & 98.26
  & 81.79
  \\
  \Disco (S\(^{\rm A6}\))       & 66.96
  & \multicolumn{1}{c|}{77.45}
  & \multicolumn{1}{c|}{86.93}
  & 88.82
  & \multicolumn{1}{c|}{59.10}
  & \multicolumn{1}{c|}{66.58}
  & 69.75
  & \multicolumn{1}{c|}{98.57}
  & \multicolumn{1}{c|}{98.61}
  & 98.73
  & 82.73
  \\
  \Disco (S\(^{\rm B6}\))       & 66.96
  & \multicolumn{1}{c|}{74.38}
  & \multicolumn{1}{c|}{86.39}
  & 88.70
  & \multicolumn{1}{c|}{57.62}
  & \multicolumn{1}{c|}{64.04}
  & 69.57
  & \multicolumn{1}{c|}{98.50}
  & \multicolumn{1}{c|}{98.45}
  & 98.57
  & 82.02
  \\
  \midrule
  TinyBERT\(^{4}\)               & 14.35                     & \multicolumn{1}{c|}{69.67}
  & \multicolumn{1}{c|}{78.35}
  & 85.12
  & \multicolumn{1}{c|}{42.66}
  & \multicolumn{1}{c|}{53.63}
  & 61.89
  & \multicolumn{1}{c|}{89.65}
  & \multicolumn{1}{c|}{96.88}
  & 97.58
  & 75.05
  \\
  UDA\(\rm_{TinyBERT^{4}}\)
  & 14.35
  & \multicolumn{1}{c|}{69.60}
  & \multicolumn{1}{c|}{77.56}
  & 83.60
  & \multicolumn{1}{c|}{40.69}
  & \multicolumn{1}{c|}{55.43}
  & 63.34
  & \multicolumn{1}{c|}{88.50}
  & \multicolumn{1}{c|}{93.63}
  & 95.98
  & 74.26
  \\
  \Disco (S\(^{\rm A4}\))
  & 14.35
  & \multicolumn{1}{c|}{77.36}
  & \multicolumn{1}{c|}{85.55}
  & 87.95
  & \multicolumn{1}{c|}{51.31}
  & \multicolumn{1}{c|}{62.93}
  &  68.24
  & \multicolumn{1}{c|}{94.79}
  & \multicolumn{1}{c|}{98.14}
  & 98.63
  & 80.54
  \\
  \Disco (S\(^{\rm B4}\))                
  & 14.35
  & \multicolumn{1}{c|}{76.90}
  & \multicolumn{1}{c|}{85.39}
  &  87.82
  & \multicolumn{1}{c|}{51.48}
  & \multicolumn{1}{c|}{62.36}
  &  68.10
  & \multicolumn{1}{c|}{94.02}
  & \multicolumn{1}{c|}{98.13}
  &  98.56
  & 80.31
  \\
   PS-NET (S\(^{\rm A4}\)) 
  & 14.35
  & \multicolumn{1}{c|}{\textbf{81.03}}
  & \multicolumn{1}{c|}{\textbf{87.32}}
  & \textbf{89.04}
  & \multicolumn{1}{c|}{\textbf{62.33}}
  & \multicolumn{1}{c|}{\textbf{68.10}}
  &  \textbf{71.35}
  & \multicolumn{1}{c|}{\textbf{97.19}}
  & \multicolumn{1}{c|}{\textbf{98.70}}
  & \textbf{98.90}
  & \textbf{\textcolor{blue}{83.77}}  \\
  PS-NET (S\(^{\rm B4}\))
  & 14.35
  & \multicolumn{1}{c|}{\textbf{82.06}}
  & \multicolumn{1}{c|}{\textbf{87.38}}
  & \textbf{89.77}
  & \multicolumn{1}{c|}{\textbf{65.21}}
  & \multicolumn{1}{c|}{\textbf{68.02}}
  &  \textbf{71.08}
  & \multicolumn{1}{c|}{\textbf{98.44}}
  & \multicolumn{1}{c|}{\textbf{98.71}}
  & \textbf{98.82}
  & \textbf{\textcolor{blue}{84.39}}  \\
  \midrule
  FLiText               & 9.60                    & \multicolumn{1}{c|}{67.14}
  & \multicolumn{1}{c|}{77.12}
  & 82.12
  & \multicolumn{1}{c|}{48.30}
  & \multicolumn{1}{c|}{57.01}
  & 63.09
  & \multicolumn{1}{c|}{89.26}
  & \multicolumn{1}{c|}{94.04}
  & 97.01
  & 75.01\\
    \Disco (S\(^{\rm A2}\))  & 8.90
  & \multicolumn{1}{c|}{75.05}
  & \multicolumn{1}{c|}{82.16}
  & 86.38
  & \multicolumn{1}{c|}{51.05}
  & \multicolumn{1}{c|}{58.83}
  & 65.63
  & \multicolumn{1}{c|}{89.55}
  & \multicolumn{1}{c|}{96.14}
  & 97.70
  & 78.05 \\
    \Disco (S\(^{\rm B2}\))               & 8.90
  & \multicolumn{1}{c|}{70.61}
  & \multicolumn{1}{c|}{81.87}
  &  86.08
  & \multicolumn{1}{c|}{48.41}
  & \multicolumn{1}{c|}{57.84}
  &  64.04
  & \multicolumn{1}{c|}{89.67}
  & \multicolumn{1}{c|}{96.06}
  &  97.58
  & 76.90
  \\ 
  PS-NET (S\(^{\rm A2}\)) 
  & 8.90
  & \multicolumn{1}{c|}{\textbf{81.14}}
  & \multicolumn{1}{c|}{\textbf{85.35}}
  & \textbf{87.10}
  & \multicolumn{1}{c|}{\textbf{61.12}}
  & \multicolumn{1}{c|}{\textbf{64.40}}
  &  \textbf{66.33}
  & \multicolumn{1}{c|}{\textbf{96.61}}
  & \multicolumn{1}{c|}{\textbf{98.24}}
  & \textbf{98.33}
  & \textbf{\textcolor{blue}{82.07}}  \\
  PS-NET (S\(^{\rm B2}\)) 
  & 8.90
  & \multicolumn{1}{c|}{\textbf{81.89}}
  & \multicolumn{1}{c|}{\textbf{87.69}}
  & \textbf{89.11}
  & \multicolumn{1}{c|}{\textbf{64.16}}
  & \multicolumn{1}{c|}{\textbf{66.88}}
  &  \textbf{69.57}
  & \multicolumn{1}{c|}{\textbf{98.05}}
  & \multicolumn{1}{c|}{\textbf{98.77}}
  & \textbf{98.57}
  & \textbf{\textcolor{blue}{83.85}}  \\
  \bottomrule
  \end{tabular}
  }
  \caption{Test accuracy (Acc (\%)) for semi-supervised text classification tasks and the baseline results are derived from \Disco. \(\mathcal{P}\)(M)  is the number of model parameters in millions. The BERT\(_{\rm BASE}\) and TinyBERT are supervised frameworks, UDA and UDA\(_{\rm TinyBERT}\), \Disco, and FLiText are semi-supervised frameworks using the same amount of unlabeled data as used by our PS-NET. The best results are in-bold, and the best average results are in-blue. } 
  \label{main_text_classification}
\end{table*}

\subsection{Baseline Methods}
For text classification, we compare with: (\(i\)) supervised baselines, BERT\(\rm_{BASE}\) and default TinyBERT~\cite{JiaoYSJCL0L20}, (\(ii\)) semi-supervised UDA~\cite{XieDHL020}, and we introduce two noteworthy lightweight semi-supervised baseline models: FLiText~\cite{LiuZFHL21}  and \Disco~\cite{JiangMLLDYW23}.  FLiText is a lightweight and fast semi-supervised learning framework and consists of a two-stage training process where it initially trains a large inspirer model (BERT) and then optimizes a target network (TextCNN). 
\Disco is the state-of-the-art faster and lighter SSL framework which employs a co-training technique to optimize multiple small student models, promoting knowledge sharing among students through diverse data and model views. We also compare with other prominent SSL text classification methods and report their results on the Unified SSL Benchmark (USB)~\cite{WangCFSTHWY0GQW22}. 
Most of these SSL methods work well on CV tasks, and ~\citet{WangCFSTHWY0GQW22} generalize them to NLP tasks by integrating a 12-layer BERT. 

For extractive summarization, we use the open-source releases: (\(i\))  supervised baseline, BERTSUM~\cite{LiuL19},  (\(ii\)) unsupervised techniques, LEAD-3, TextRank~\cite{MihalceaT04} and LexRank~\cite{ErkanR04} and  (\(iii\)) two state-of-the-art semi-supervised extractive summarization methods, UDASUM and CPSUM~\cite{WangMLJZL22} for comparison.


\begin{table}[tb]
  \centering
  \footnotesize
 \caption{ROUGE F1 performance of the extractive summarization. L\(_{d}\)=100 refers to the labelled samples. 
  SSL baselines (CPSUM, UDASUM, \Disco) use the same unlabeled data as our PS-NET.}
  \renewcommand\arraystretch{1.2}
  \setlength{\abovecaptionskip}{1.1mm}
  \setlength{\tabcolsep}{1.75mm}{
  \begin{tabular}{@{}llc|ccr@{}}
  \toprule
  \multirow{2}{*}{\textbf{Models}} & \multirow{2}{*}{\textbf{\(\mathcal{P}\)(M)}} & \multirow{2}{*}{\textbf{L\(_{d}\)}} & \multicolumn{3}{c}{\textbf{CNN/DailyMail}}                                      \\ \cline{4-6}
  &                         &            & \multicolumn{1}{l}{R-1} & \multicolumn{1}{c}{R-2} & \multicolumn{1}{r@{}}{R-L}  \\
  \midrule
  ORACLE    & \multirow{4}{*}{} & 100   & \emph{48.35}   & \emph{26.28}  & \emph{44.61}  \\
  \midrule
  LEAD-3    &                   & 100   & 40.04   & 17.21  & 36.14  \\
  TextRank   &                   & 100   & 33.84   & 13.11  & 23.98  \\
  LexRank    &                   & 100   & 34.63   & 12.72  & 21.25  \\
  \midrule
  BERTSUM & 109.48                 & 100   & 38.58   & 15.97  & 34.79  \\
  CPSUM    & 109.48                 & 100   & 38.10   & 15.90  & 34.39  \\
  UDASUM   & 109.48                  & 100   & 38.58   & 15.87  & 34.78  \\
  TinyBERTSUM\(^{4}\)   & 14.35                 & 100   & 39.83   & 17.24  & 35.98  \\
  UDASUM\(\rm_{TinyBERT^{4}}\)     & 14.35                & 100   & 40.11   & 17.43  & 36.23  \\
  \Disco (S\(\rm^{A4}\))      & 14.35                & 100   & 40.39  & 17.55 & 36.47  \\
  \Disco (S\(\rm^{B4}\))      & 14.35                 & 100   & 40.40   & 17.57  & 36.48  \\   \midrule
  PS-NET (S\(\rm^{A4}\))      & 14.35                & 100   & \textbf{41.29}   & \textbf{17.78}  & \textbf{37.11}  \\
  PS-NET (S\(\rm^{B4}\))      & 14.35                 & 100   & \textbf{40.74}   & \textbf{17.86}  & \textbf{37.16}  \\  \bottomrule
  \end{tabular}
  }
  \label{text_summarization_1}
\end{table}

\section{Experimental Results}
\subsection{Evaluation on Text Classification}


Upon comparing the 2-layer students of PS-NET with the 12-layer BERT, our method demonstrates a notable performance enhancement across various text classification tasks, despite a 12.3\(\times\) reduction in model size. 
The 2-layer students of PS-NET notably surpass the 4-layer and 6-layer semi-supervised UDA\(_{\rm TinyBERT}\) and \Disco by a margin in semi-supervised text classification. With minimal labelled data, our PS-NET (the inferior student) featuring a 4-layer distilled BERT, outperforms the 4-layer UDA\(_{\rm TinyBERT}\) by an average margin of 9.51\% across three datasets.
Notably, PS-NET demonstrates robust performance, even with a minimal annotated data size of 10 per class. 
The superior student in PS-NET, equipped with a 2-layer distilled BERT, exhibits a substantial average performance improvement of 8.84\% over FLiText and outperforms the best student of 2-layer \Disco by 5.80\% across three datasets.

\begin{table}[t]
  \centering
  \footnotesize
  \caption{Test accuracy (Acc (\%)) of other prominent SSL text classification models in Amazon Review and Yelp Review datasets. Results of baselines are sourced from the most up-to-date results on GitHub of USB Benchmark~\cite{WangCFSTHWY0GQW22}. Results underlined indicate performance inferior to PS-NET.  Svp refers to supervised methods. 
  }
  \renewcommand\arraystretch{1.2}
  \setlength{\abovecaptionskip}{2mm}
  \setlength{\tabcolsep}{.26mm}{
  \begin{tabular}{@{}l|c|c|c|c|r@{}}
    \toprule
  \textbf{Models} & \textbf{\(\mathcal{P}\)(M)}      & \textbf{Amzn}-{\tiny 50} & \textbf{Amzn}-{\tiny 200} & \textbf{Yelp}-{\tiny 50} & \textbf{Yelp}-{\tiny 200}\\ \hline
  Fully-Svp	     
  &  \cellcolor{gray!30}109.48   & 63.60  &63.60  &67.96   &67.96  \\
  Svp 
  &  \cellcolor{gray!30}109.48   & 48.69  &\underline{52.47}  &\underline{49.78}   &\underline{53.29}  \\
  P-Labeling
  &  \cellcolor{gray!30}109.48   & \underline{46.55}  &\underline{52.34}  &\underline{49.40}   &\underline{52.79}  \\
  \(\prod \)-model
  &  \cellcolor{gray!30}109.48   & \underline{22.78}  &\underline{46.83}  &\underline{24.27}   &\underline{40.18}  \\
  MeanTch
  &  \cellcolor{gray!30}109.48   & 47.86  &\underline{52.34}  &\underline{49.40}   &\underline{52.79}  \\
  VAT
  & \cellcolor{gray!30}109.48   & 50.17  &\underline{53.46} & \underline{47.03}   &\underline{54.70}  \\
  UDA
  &  \cellcolor{gray!30}109.48   & \underline{39.24}  &\underline{31.62} &\underline{30.67}   &\underline{33.05}  \\
  FixMatch
  &  \cellcolor{gray!30}109.48   & 52.39   &  56.95  &  53.48  &   59.35   \\
  Flexmatch
  &  \cellcolor{gray!30}109.48   & 54.27   &  57.75  &  56.65  &   59.49   \\
  AdaMatch
  &  \cellcolor{gray!30}109.48   & 53.28   &  57.73  &  54.60  &   59.84   \\
  SimMatch
  &  \cellcolor{gray!30}109.48   & 54.09   &  57.79  &  53.88  &   59.74   \\
  FreeMatch
  &  \cellcolor{gray!30}109.48   & 53.59   &  57.36  &  52.05  &   59.63   \\
  SoftMatch
  &  \cellcolor{gray!30}109.48   & 54.71   &  57.79  &  55.91  &   60.24   \\
  \Disco  (S\(\rm^{A4}\))
  &  \cellcolor{gray!10}14.35   & \underline{46.28}   &  \underline{48.64}  &  \underline{45.42}  &   \underline{50.87}   \\
  \Disco  (S\(\rm^{B4}\))
  &  \cellcolor{gray!10}14.35   & \underline{36.51}   &  \underline{46.41}  &  \underline{38.47}  &   \underline{49.25}   \\
  PS-NET  (S\(\rm^{A4}\))
  &  \cellcolor{gray!10}14.35   & 46.22   &  53.23  &  49.68  &   55.32   \\
  PS-NET  (S\(\rm^{B4}\))
  &  \cellcolor{gray!10}14.35   & \textbf{47.56}   &  \textbf{54.77}  &  \textbf{50.12}  &   \textbf{57.65}   \\
        \bottomrule
  \end{tabular}
  }
  \label{auxiliary_text_classification_1}
\end{table}

\begin{table}[tb]
  \footnotesize
 \caption{Validation accuracy (Acc (\%)) of PS-NET with multiple student peers. The labelled data consists of 10 samples per class. The students (S\(^{\rm A2}\), S\(^{\rm B2}\), S\(^{\rm C2}\), S\(^{\rm D2}\), S\(^{\rm E2}\), S\(^{\rm F2}\))  are distilled from layers \(\left \{ 1,2 \right \}\),  \(\left \{11,12 \right \}\), \(\left \{3,4\right \}\),  \(\left \{5,6\right \}\), \(\left \{7,8\right \}\), \(\left \{9,10\right \}\) of the teacher BERT\(_{\rm BASE}\). The students (S\(^{\rm A6}\), S\(^{\rm B6}\))  are distilled from layers \(\left \{ 1,2,3,4,5,6 \right \}\),   \(\left \{7,8,9,10,11,12\right \}\) of BERT\(_{\rm BASE}\). 
  }
        \renewcommand\arraystretch{1.2}
     \setlength{\abovecaptionskip}{0.5mm}
     \setlength{\tabcolsep}{2.4mm}{
  \begin{tabular}{@{}l|c|c|r@{}}
    \toprule
  \textbf{Models}                            &  \textbf{AG News} & \textbf{Yahoo!Answer} & \textbf{DBpedia} \\ 
  \hline
  PS-NET (S\(^{\rm A2}\))     & 81.14    & 61.12                           & 96.61       \\  
  PS-NET (S\(^{\rm B2}\))         & 81.89    &  63.91                          & 98.05       \\ \cline{1-2} \cline{3-4} 
  PS-NET (S\(^{\rm A2}\))    & 81.20      &     60.22                         & 95.70       \\  
   PS-NET (S\(^{\rm B2}\))     &  \textbf{83.50}     &  64.14                       & 98.43      \\
  PS-NET (S\(^{\rm C2}\))      & 81.33      &  61.05                         & 96.43     \\  
  PS-NET (S\(^{\rm D2}\))                     & 81.45      &    61.11                        &  96.87      \\  
  PS-NET (S\(^{\rm E2}\))        & 
  82.41      &    61.28                          & 97.55        \\  
  PS-NET (S\(^{\rm F2}\))       & 81.34       &       62.19                    &  98.21       \\  
  \cline{1-2} \cline{3-4}
  PS-NET (S\(^{\rm A6}\))    & 82.39    & 64.12                           & 98.51       \\  
  PS-NET (S\(^{\rm B6}\))      & 82.61    &  \textbf{64.33}                         & \textbf{98.49}      \\  
  \bottomrule
  \end{tabular}
     }
     \label{multi_students}
\end{table}

Table~\ref{auxiliary_text_classification_1} provides a comparative analysis of PS-NET with other notable SSL methods equipped with a 12-layer BERT, utilizing results from the Unified SSL Benchmark (USB)~\cite{WangCFSTHWY0GQW22}. 
Remarkably, PS-NET's 4-layer BERT-based students outperform most of these methods. These findings underscore the superiority of our model in scenarios involving lightweight model architecture and limited labelled data across various text classification tasks.


\subsection{Evaluation on Extractive Summarization}
Results of the low-resource performance of the extractive summarization on CNN/DailyMail are shown in Table~\ref{text_summarization_1}. 
Our approach is obviously superior to all supervised and SSL baselines, with only 100 labelled samples available.

In the summarization task, marginal performance differences exist among the SSL models. These discrepancies can be attributed partially to the inherent difficulty of 2-class sentence classification and the collapse problem~\cite{YanLWZWX20,ChenH21,GaoYC21} associated with BERT sentence representation. These difficulties are worsened by the constraint of having only 100 labelled summaries for training extractive models. Despite these factors, our method outperforms all existing SSL models in extractive summarization tasks, indicating its suitability for scenarios characterized by severe data scarcity issues.

\begin{figure*}
  \centering
  \begin{minipage}[t]{0.24\linewidth}
  \centering
  \subfigure[SingleStudent$^{\rm A6}$]{
    \includegraphics[width=1.53in]{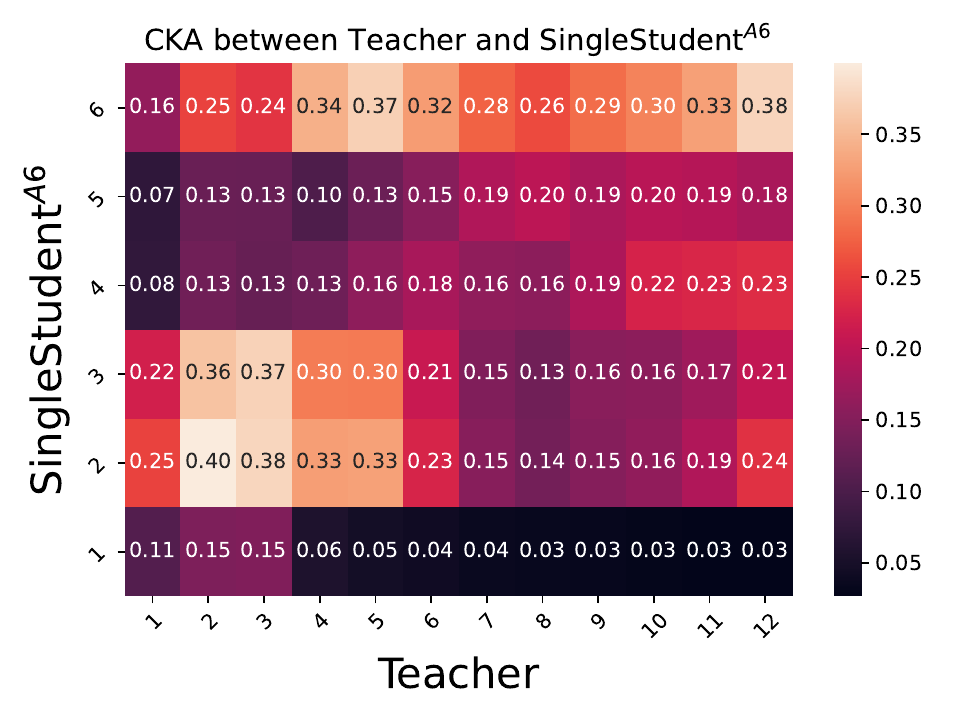}}
  \end{minipage}%
  \begin{minipage}[t]{0.24\linewidth}
  \centering
      \subfigure[SingleStudent$^{\rm B6}$]{
  \includegraphics[width=1.53in]{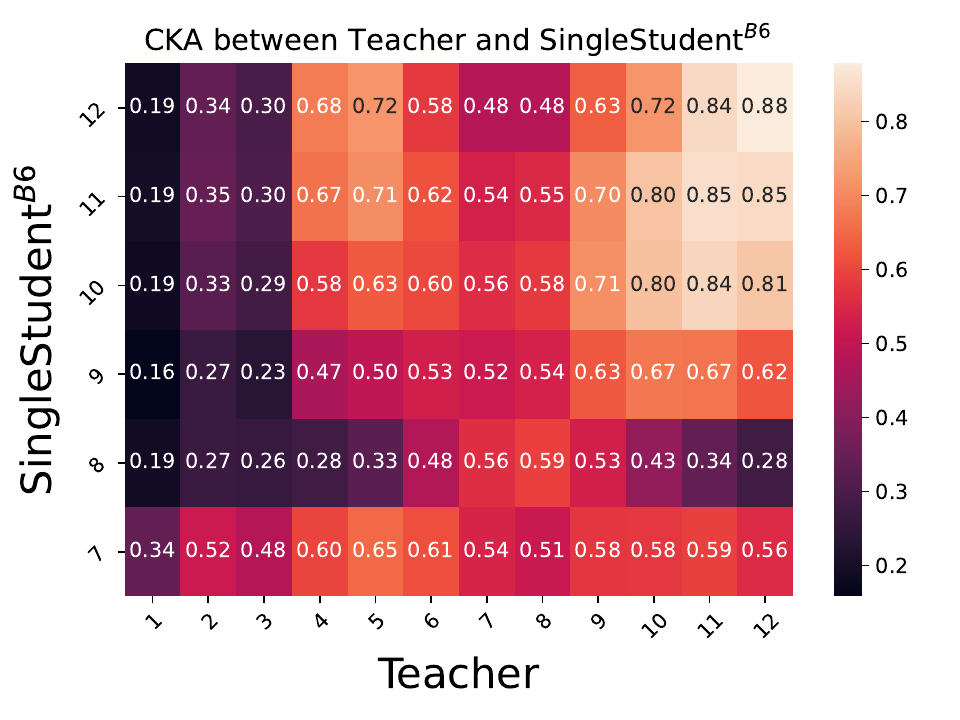}}
      \label{EGAT}
  \end{minipage}
    \begin{minipage}[t]{0.24\linewidth}
  \centering
      \subfigure[PS-NET {\scriptsize only w/ DML} $(\rm S^{A6})$]{
  \includegraphics[width=1.53in]{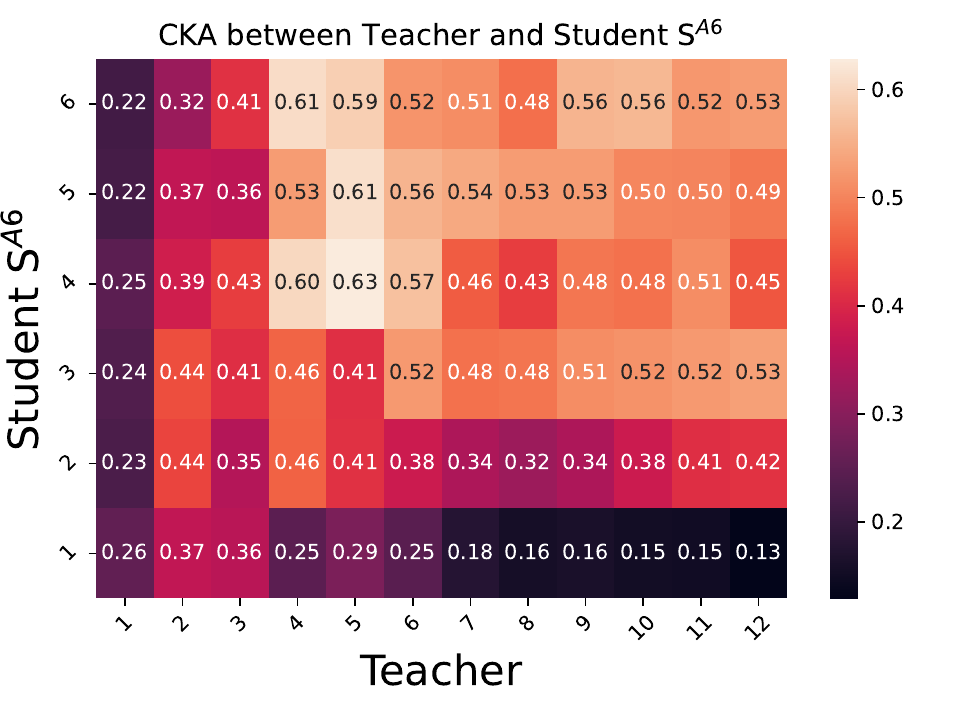}}
      \label{NGAT}
  \end{minipage}
    \begin{minipage}[t]{0.24\linewidth}
  \centering
      \subfigure[PS-NET {\scriptsize only w/ DML} $(\rm S^{B6})$]{
  \includegraphics[width=1.53in]{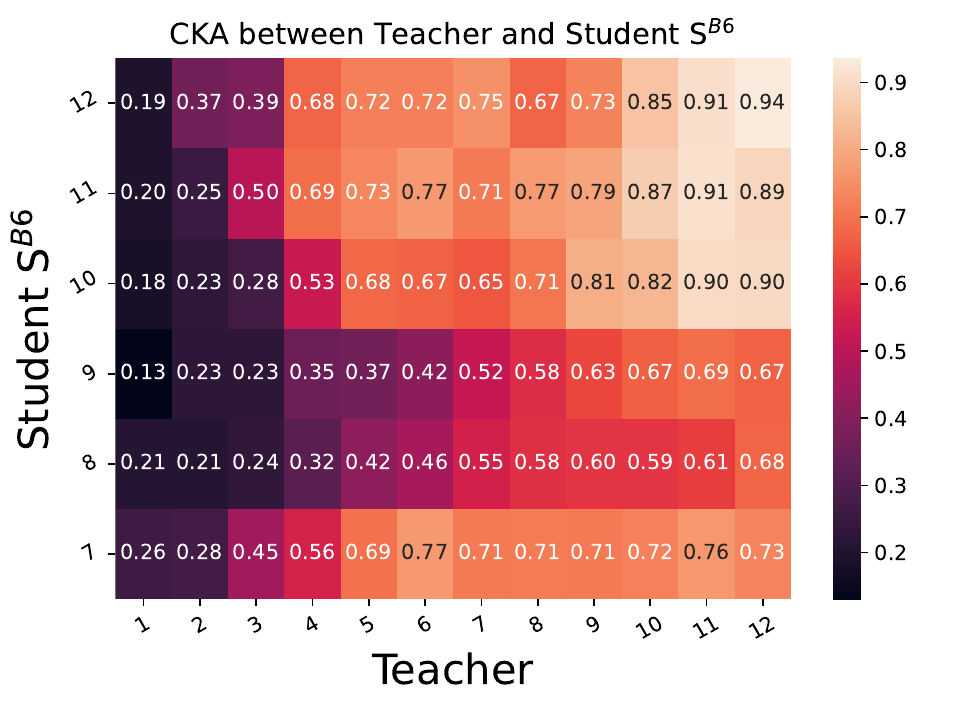}}
      \label{ENGAT}
  \end{minipage}
      \vspace{-0.1in}
     \caption{The visualization of the Center Kernel Alignment (CKA~\cite{ZhuW21a}) scores of PS-NET in Subfigures (c) and (d), along with its ablation variant, SingleStudent, shown in Subfigures (a) and (b). All models utilize a 6-layer BERT  on the AG News dataset for evaluation, with 10 labelled examples per class.}
    \label{loss_2D}
\end{figure*}

\subsection{Qualitative Analysis}
In this section, we investigate the impacts of DML \& CAT, and the scaling of students and layers within our PS-NET framework. Model efficiency is detailed in Appendix~\ref{Model_Efficiency}.

\noindent\textbf{Benefits of Mutual-learning (Peer Collaboration)}. 
To compare the advantages of mutual learning within our framework, we establish a SingleStudent setup with one student, employing the distillation loss as a regularization term alongside the supervised loss. It is crucial to note that the SingleStudent model differs from TinyBERT\footnote{First, the SingleStudent uses online distillation with both labelled and unlabeled data in our two-stage framework, while TinyBERT begins with unlabeled data for general distillation and fine-tunes with labelled data. Second, TinyBERT separates labelled and unlabeled training, where the teacher optimizes labels while distilling knowledge.}.


As shown in Table~\ref{single_student}, under the two-stage framework of supervised knowledge optimization and unsupervised model compression, the performance of the 
SingleStudent is notably weaker compared to that of two students engaged in mutual learning. Figure~\ref{loss_2D} further supports this, showing that mutual learning improves the Center Kernel Alignment (CKA) score between students and the teacher, effectively narrowing the performance gap. This improvement is due to mutual learning in PS-NET, where students indirectly extract regularization from each other's predictive capabilities, thereby enhancing their individual generalization abilities.

\begin{table}[tb]
  \footnotesize
 \caption{Validation accuracy (Acc (\%)) comparison between PS-NET's students and the SingleStudent model under limited 10 labelled data per class. The SingleStudent undergoes a two-stage learning process with only one teacher and one student. }
        \renewcommand\arraystretch{1.2}
     \setlength{\abovecaptionskip}{0.5mm}
     \setlength{\tabcolsep}{2.mm}{
 \begin{tabular}{@{}l|c|c|r@{}}
 \toprule
 \textbf{Models}           & \textbf{AG News} & \textbf{Yahoo!Answer} & \textbf{DBpedia} \\ \hline
$\rm Single Student^{A2}$                    & 80.29          & 61.46                & 96.13   \\      
 $\rm Single Student^{B2}$                   & 80.66           & 63.10                    & 97.48    \\
 PS-NET $(\rm S^{A2})$                   & \textbf{81.14}          & 61.12                 & 96.61   \\      
 PS-NET $(\rm S^{B2})$                 & \textbf{81.89}           & \textbf{63.91}                 & \textbf{98.05}    \\
 \bottomrule
\end{tabular}
}
\label{single_student}
\end{table}

\noindent\textbf{Benefits of Curriculum Adversarial Learning (Self Transcendence)}. 
The ablation experimental results from Table~\ref{CAT_comparison} underscore the substantial performance boost of PS-NET when Curriculum Adversarial Training (CAT) is integrated into its training regimen. 
Within the FS-NET framework, guided by the CAT, one student consistently emerges in optimal performance. This improvement indicates that CAT effectively enhances the model's generalization capability.



\begin{table}[t]
  \footnotesize
  \caption{Validation accuracy (Acc (\%)) of PS-NET using the minimum labelled data, comparing with and without curriculum adversarial training (CAT).}
  \renewcommand\arraystretch{1.2}
     \setlength{\abovecaptionskip}{0.5mm}
     \setlength{\tabcolsep}{0.365mm}{
  \begin{tabular}{@{}l|c|c|c|r@{}}
    \toprule
    \textbf{Models}    & \textbf{AG News} & \textbf{Yaho..} & \textbf{Amzn}-{\tiny 50} &\textbf{DBpedia} \\ \hline
  FS-NET (S\(^{\rm A2}\)).w/o {\tiny CAT} & 80.37  & 62.29 & 44.15  & 97.65   \\ 
  FS-NET (S\(^{\rm B2}\)).w/o {\tiny CAT} & 78.29  & 61.16 & 44.74  & 97.50   \\ 
  FS-NET (S\(^{\rm A2}\)).w {\tiny CAT}   & \textbf{81.14}  & 61.12 & 43.18  & 96.61   \\ 
  FS-NET (S\(^{\rm B2}\)).w {\tiny CAT}   & \textbf{81.89}  & \textbf{63.91} & \textbf{45.31}  & \textbf{98.05}   \\ 
  \bottomrule
\end{tabular}
     }
     \label{CAT_comparison}
\end{table}

\noindent\textbf{Effect of More Student Peers}. 
The prior experiments study the dual-student cohort.
We next investigate how PS-NET scales with more students in the cohort.
In PS-NET, each student learns from all other students individually, regardless of how many students are in the cohort. 
As shown in Table~\ref{multi_students},  expanding to a four-student cohort in PS-NET enhances individual student performance, showcasing improved generalization as peer numbers increase.  
These results demonstrate that more student perturbations complement each other and are important to obtain superior performance for PS-NET with a compressed model and few labelled data.

\noindent\textbf{Effect of More Student Layers}. 
As shown in Table~\ref{multi_students}, two students with multiple layers exhibited performance advantages on the three datasets. 
It indicates that more student feature encoding diversifies the cohort and then encourages the individual models to teach each other in a complementary manner underlying multiple views to improve the learning performances.  In practical scenarios, multiple lighter-weight students remain the preferred option, as smaller individual learning parameters result in faster inference speeds.

\section{Conclusion and Future Work}

We present a novel framework, PS-NET, designed to address both label scarcity and model size reduction in pre-trained language models (PLMs) such as BERT. PS-NET incorporates lightweight student cohorts to facilitate mutual learning and adversarial training, thereby enhancing generalization. In the future, we aim to extend PS-NET to other NLP benchmarks, including language understanding, machine reading comprehension, and text generation tasks. Additionally, leveraging other language models, such as RoBERTa~\cite{abs190711692} and GPT~\cite{radford2018improving,radford2019language,BrownMRSKDNSSAA20}, will further contribute to our work.




\section{Broader Impacts \& Limitations}
PS-NET furnishes robust technical solutions for faster and lighter semi-supervised learning, providing an effective way to deploy it on resource-limited devices and industrial applications. 

A substantial distinction of PS-NET exists between the compression challenge encountered in large generative models, such as ChatGPT, and the standard knowledge distillation paradigm explored in PS-NET. 
Notably, ~\citet{abs-2308-07633} categorize standard knowledge distillation as falling within the White-box KD category, where the teacher's parameters are available for use. Conversely, the compressed Black-box KD applied to ChatGPT requires students to grasp both the teachers' knowledge and their emergent abilities~\cite{abs-2206-07682}. 
These abilities should be distilled by the teacher and be imitated by the students, including In-Context Learning (ICL)~\cite{abs-2301-00234,abs-2301-11916}, Chain of Thought (CoT)~\cite{ShiSF0SVCTRZ0W23}, and Instruction Following (IF)~\cite{Ouyang0JAWMZASR22,BrooksHE23}. While various technological paradigms and explicit workloads prompt discussions on generative large language models for future research, BERT proves more applicable and sufficient for the discriminative tasks addressed in our works.

Besides, Few-shot LLMs emphasize task-agnostic pretraining LLM and its transferability with limited samples, while we focus on small, randomly initialized models tailored for limited labeled data. Recent few-shot methods like ICL~\cite{abs-2301-00234,abs-2301-11916}, PEFT~\cite{abs-2403-13372}, and prompt-fre SetFit achieve strong results with minimal labeled data and without relying on unlabeled data, yet rely heavily on the pretrained performance of large models. Applying such methods to our PS-NET, which relies on arbitrarily small, randomly initialized models, poses significant challenges and expected workload.

Furthermore, distillation with LLMs is a meaningful aspect of evaluating model PS-NET scalability. However, LLMs encounter  challenges in serving as teachers to train significantly smaller student models. Previous studies have discussed the inferior performance of knowledge distillation when there is a significant size disparity between the teacher and student models~\cite{ChoH19,MirzadehFLLMG20,LiFSLLYZ22,0020Y00022}. Existing research on LLM-based knowledge distillation primarily focuses on extracting and transferring the rich, nuanced understanding developed by these models rather than merely reducing the model size. Additionally, while ensembling models~\cite{ChenG16,KeMFWCMYL17} boosts performance (as demonstrated by \citet{JiangMLLDYW23}) but contradicts our goal of a singular model for faster inference. 
We focus on training a lightweight model with limited labels for efficient inference. 

Therefore, several promising avenues for further exploration exist within the PS-NET framework, with numerous optimization opportunities.

\section{Ethical Statement}
  \noindent\textbf{Data Availability and Safety.} The experimental data analyzed in this paper are primarily publicly accessible; otherwise, we will provide links upon request for access. 

  \noindent\textbf{Usage of Large PLM.} The paper does not engage in additional utilization of the GPT-3.5 model or its derivatives, such as generating content for manuscripts or coding for models.

  \noindent\textbf{Authors Conflicts.}
  The authors assert that they have no conflicts of interest. Informed consent has been obtained from all individual participants involved in the study. There are no financial or non-financial relationships or interests that could be perceived as potentially influencing the publication.

\bibliography{acl_latex}

\begin{thebibliography}{71}
\providecommand{\natexlab}[1]{#1}

\bibitem[{Berthelot et~al.(2019)Berthelot, Carlini, Goodfellow, Papernot, Oliver, and Raffel}]{BerthelotCGPOR19}
David Berthelot, Nicholas Carlini, Ian~J. Goodfellow, Nicolas Papernot, Avital Oliver, and Colin Raffel. 2019.
\newblock \href {https://proceedings.neurips.cc/paper/2019/hash/1cd138d0499a68f4bb72bee04bbec2d7-Abstract.html} {Mixmatch: {A} holistic approach to semi-supervised learning}.
\newblock In \emph{{NeurIPS}}, pages 5050--5060.

\bibitem[{Berthelot et~al.(2022)Berthelot, Roelofs, Sohn, Carlini, and Kurakin}]{BerthelotRSCK22}
David Berthelot, Rebecca Roelofs, Kihyuk Sohn, Nicholas Carlini, and Alexey Kurakin. 2022.
\newblock \href {https://openreview.net/forum?id=Q5uh1Nvv5dm} {Adamatch: {A} unified approach to semi-supervised learning and domain adaptation}.
\newblock In \emph{{ICLR}}. OpenReview.net.

\bibitem[{Board and Pitt(1989)}]{BoardP89}
Raymond~A. Board and Leonard Pitt. 1989.
\newblock \href {https://doi.org/10.1007/BF00114803} {Semi-supervised learning}.
\newblock \emph{Mach. Learn.}, 4:41--65.

\bibitem[{Brooks et~al.(2023)Brooks, Holynski, and Efros}]{BrooksHE23}
Tim Brooks, Aleksander Holynski, and Alexei~A. Efros. 2023.
\newblock \href {https://doi.org/10.1109/CVPR52729.2023.01764} {Instructpix2pix: Learning to follow image editing instructions}.
\newblock In \emph{{CVPR}}, pages 18392--18402. {IEEE}.

\bibitem[{Brown et~al.(2020)Brown, Mann, Ryder, Subbiah, Kaplan, Dhariwal, Neelakantan, Shyam, Sastry, Askell, Agarwal, Herbert{-}Voss, Krueger, Henighan, Child, Ramesh, Ziegler, Wu, Winter, Hesse, Chen, Sigler, Litwin, Gray, Chess, Clark, Berner, McCandlish, Radford, Sutskever, and Amodei}]{BrownMRSKDNSSAA20}
Tom~B. Brown, Benjamin Mann, Nick Ryder, Melanie Subbiah, Jared Kaplan, Prafulla Dhariwal, Arvind Neelakantan, Pranav Shyam, Girish Sastry, Amanda Askell, Sandhini Agarwal, Ariel Herbert{-}Voss, Gretchen Krueger, Tom Henighan, Rewon Child, Aditya Ramesh, Daniel~M. Ziegler, Jeffrey Wu, Clemens Winter, Christopher Hesse, Mark Chen, Eric Sigler, Mateusz Litwin, Scott Gray, Benjamin Chess, Jack Clark, Christopher Berner, Sam McCandlish, Alec Radford, Ilya Sutskever, and Dario Amodei. 2020.
\newblock \href {https://proceedings.neurips.cc/paper/2020/hash/1457c0d6bfcb4967418bfb8ac142f64a-Abstract.html} {Language models are few-shot learners}.
\newblock In \emph{NeurIPS}.

\bibitem[{Chang et~al.(2008)Chang, Ratinov, Roth, and Srikumar}]{ChangRRS08}
Ming{-}Wei Chang, Lev{-}Arie Ratinov, Dan Roth, and Vivek Srikumar. 2008.
\newblock \href {http://www.aaai.org/Library/AAAI/2008/aaai08-132.php} {Importance of semantic representation: Dataless classification}.
\newblock In \emph{{AAAI}}, pages 830--835. {AAAI} Press.

\bibitem[{Chen et~al.(2023)Chen, Tao, Fan, Wang, Wang, Schiele, Xie, Raj, and Savvides}]{abs-2301-10921}
Hao Chen, Ran Tao, Yue Fan, Yidong Wang, Jindong Wang, Bernt Schiele, Xing Xie, Bhiksha Raj, and Marios Savvides. 2023.
\newblock \href {https://doi.org/10.48550/arXiv.2301.10921} {Softmatch: Addressing the quantity-quality trade-off in semi-supervised learning}.
\newblock \emph{CoRR}, abs/2301.10921.

\bibitem[{Chen et~al.(2020)Chen, Yang, and Yang}]{ChenYY20}
Jiaao Chen, Zichao Yang, and Diyi Yang. 2020.
\newblock \href {https://doi.org/10.18653/v1/2020.acl-main.194} {Mixtext: Linguistically-informed interpolation of hidden space for semi-supervised text classification}.
\newblock In \emph{{ACL}}, pages 2147--2157. Association for Computational Linguistics.

\bibitem[{Chen et~al.(2024)Chen, Zhang, Chen, and Hu}]{ChenZCH24}
Junfan Chen, Richong Zhang, Junchi Chen, and Chunming Hu. 2024.
\newblock \href {https://doi.org/10.18653/V1/2024.ACL-LONG.118} {Open-set semi-supervised text classification via adversarial disagreement maximization}.
\newblock In \emph{{ACL}}, pages 2170--2180. Association for Computational Linguistics.

\bibitem[{Chen and Guestrin(2016)}]{ChenG16}
Tianqi Chen and Carlos Guestrin. 2016.
\newblock \href {https://doi.org/10.1145/2939672.2939785} {Xgboost: {A} scalable tree boosting system}.
\newblock In \emph{SIGKDD}, pages 785--794. {ACM}.

\bibitem[{Chen and He(2021)}]{ChenH21}
Xinlei Chen and Kaiming He. 2021.
\newblock \href {https://openaccess.thecvf.com/content/CVPR2021/html/Chen\_Exploring\_Simple\_Siamese\_Representation\_Learning\_CVPR\_2021\_paper.html} {Exploring simple siamese representation learning}.
\newblock In \emph{{CVPR}}, pages 15750--15758. Computer Vision Foundation / {IEEE}.

\bibitem[{Cho and Hariharan(2019)}]{ChoH19}
Jang~Hyun Cho and Bharath Hariharan. 2019.
\newblock On the efficacy of knowledge distillation.
\newblock In \emph{{ICCV}}, pages 4793--4801. {IEEE}.

\bibitem[{Devlin et~al.(2019)Devlin, Chang, Lee, and Toutanova}]{DevlinCLT19}
Jacob Devlin, Ming{-}Wei Chang, Kenton Lee, and Kristina Toutanova. 2019.
\newblock \href {https://doi.org/10.18653/v1/n19-1423} {{BERT:} pre-training of deep bidirectional transformers for language understanding}.
\newblock In \emph{{NAACL-HLT}}, pages 4171--4186. Association for Computational Linguistics.

\bibitem[{Dong et~al.(2023)Dong, Li, Dai, Zheng, Wu, Chang, Sun, Xu, Li, and Sui}]{abs-2301-00234}
Qingxiu Dong, Lei Li, Damai Dai, Ce~Zheng, Zhiyong Wu, Baobao Chang, Xu~Sun, Jingjing Xu, Lei Li, and Zhifang Sui. 2023.
\newblock \href {https://doi.org/10.48550/ARXIV.2301.00234} {A survey for in-context learning}.
\newblock \emph{CoRR}, abs/2301.00234.

\bibitem[{Erkan and Radev(2004)}]{ErkanR04}
G{\"{u}}nes Erkan and Dragomir~R. Radev. 2004.
\newblock \href {https://doi.org/10.1613/jair.1523} {Lexrank: Graph-based lexical centrality as salience in text summarization}.
\newblock \emph{J. Artif. Intell. Res.}, 22:457--479.

\bibitem[{Fan et~al.(2023)Fan, Kukleva, Dai, and Schiele}]{DBLP:journals/ijcv/FanKDS23/revisitconsistency}
Yue Fan, Anna Kukleva, Dengxin Dai, and Bernt Schiele. 2023.
\newblock \href {https://doi.org/10.1007/s11263-022-01723-4} {Revisiting consistency regularization for semi-supervised learning}.
\newblock \emph{Int. J. Comput. Vis.}, 131(3):626--643.

\bibitem[{Furlanello et~al.(2018)Furlanello, Lipton, Tschannen, Itti, and Anandkumar}]{FurlanelloLTIA18}
Tommaso Furlanello, Zachary~Chase Lipton, Michael Tschannen, Laurent Itti, and Anima Anandkumar. 2018.
\newblock Born-again neural networks.
\newblock In \emph{{ICML}}, volume~80 of \emph{Proceedings of Machine Learning Research}, pages 1602--1611. {PMLR}.

\bibitem[{Gao et~al.(2021)Gao, Yao, and Chen}]{GaoYC21}
Tianyu Gao, Xingcheng Yao, and Danqi Chen. 2021.
\newblock \href {https://doi.org/10.18653/v1/2021.emnlp-main.552} {Simcse: Simple contrastive learning of sentence embeddings}.
\newblock In \emph{{EMNLP}}, pages 6894--6910. Association for Computational Linguistics.

\bibitem[{Guo et~al.(2023)Guo, Ma, Su, Wang, Zhao, Zou, Sun, and Zheng}]{GuoMSWZZSZ23}
Yuxin Guo, Shijie Ma, Hu~Su, Zhiqing Wang, Yuhao Zhao, Wei Zou, Siyang Sun, and Yun Zheng. 2023.
\newblock \href {http://papers.nips.cc/paper\_files/paper/2023/hash/98143953a7fd1319175b491888fc8df5-Abstract-Conference.html} {Dual mean-teacher: An unbiased semi-supervised framework for audio-visual source localization}.
\newblock In \emph{NeurIPS}.

\bibitem[{He et~al.(2024)He, Su, Cai, Shalyminov, Song, and Mansour}]{HeSCSSM24}
Jianfeng He, Hang Su, Jason Cai, Igor Shalyminov, Hwanjun Song, and Saab Mansour. 2024.
\newblock \href {https://doi.org/10.18653/V1/2024.NAACL-LONG.333} {Semi-supervised dialogue abstractive summarization via high-quality pseudolabel selection}.
\newblock In \emph{{NAACL}}, pages 5976--5996. Association for Computational Linguistics.

\bibitem[{Hermann et~al.(2015)Hermann, Kocisk{\'{y}}, Grefenstette, Espeholt, Kay, Suleyman, and Blunsom}]{HermannKGEKSB15}
Karl~Moritz Hermann, Tom{\'{a}}s Kocisk{\'{y}}, Edward Grefenstette, Lasse Espeholt, Will Kay, Mustafa Suleyman, and Phil Blunsom. 2015.
\newblock \href {https://proceedings.neurips.cc/paper/2015/hash/afdec7005cc9f14302cd0474fd0f3c96-Abstract.html} {Teaching machines to read and comprehend}.
\newblock In \emph{{NeurIPS}}, pages 1693--1701.

\bibitem[{Hinton et~al.(2012)Hinton, Srivastava, Krizhevsky, Sutskever, and Salakhutdinov}]{hinton2012improving}
Geoffrey~E Hinton, Nitish Srivastava, Alex Krizhevsky, Ilya Sutskever, and Ruslan~R Salakhutdinov. 2012.
\newblock \href {http://arxiv.org/abs/1207.0580} {Improving neural networks by preventing co-adaptation of feature detectors}.
\newblock \emph{arXiv preprint arXiv:1207.0580}.

\bibitem[{Huang et~al.(2022)Huang, You, Wang, Qian, and Xu}]{0020Y00022}
Tao Huang, Shan You, Fei Wang, Chen Qian, and Chang Xu. 2022.
\newblock Knowledge distillation from {A} stronger teacher.
\newblock In \emph{NeurIPS}.

\bibitem[{Jiang et~al.(2023)Jiang, Mao, Lin, Li, Deng, Yang, and Wang}]{JiangMLLDYW23}
Weifeng Jiang, Qianren Mao, Chenghua Lin, Jianxin Li, Ting Deng, Weiyi Yang, and Zheng Wang. 2023.
\newblock \href {https://aclanthology.org/2023.emnlp-main.244} {Disco: Distilled student models co-training for semi-supervised text mining}.
\newblock In \emph{{EMNLP}}, pages 4015--4030. Association for Computational Linguistics.

\bibitem[{Jiao et~al.(2020)Jiao, Yin, Shang, Jiang, Chen, Li, Wang, and Liu}]{JiaoYSJCL0L20}
Xiaoqi Jiao, Yichun Yin, Lifeng Shang, Xin Jiang, Xiao Chen, Linlin Li, Fang Wang, and Qun Liu. 2020.
\newblock \href {https://doi.org/10.18653/v1/2020.findings-emnlp.372} {Tinybert: Distilling {BERT} for natural language understanding}.
\newblock In \emph{{EMNLP}}, volume {EMNLP} 2020 of \emph{Findings of {ACL}}, pages 4163--4174. Association for Computational Linguistics.

\bibitem[{Ke et~al.(2017)Ke, Meng, Finley, Wang, Chen, Ma, Ye, and Liu}]{KeMFWCMYL17}
Guolin Ke, Qi~Meng, Thomas Finley, Taifeng Wang, Wei Chen, Weidong Ma, Qiwei Ye, and Tie{-}Yan Liu. 2017.
\newblock \href {https://proceedings.neurips.cc/paper/2017/hash/6449f44a102fde848669bdd9eb6b76fa-Abstract.html} {Lightgbm: {A} highly efficient gradient boosting decision tree}.
\newblock In \emph{NeuralPS}, pages 3146--3154.

\bibitem[{Ke et~al.(2019)Ke, Wang, Yan, Ren, and Lau}]{KeWYRL19}
Zhanghan Ke, Daoye Wang, Qiong Yan, Jimmy S.~J. Ren, and Rynson W.~H. Lau. 2019.
\newblock \href {https://doi.org/10.1109/ICCV.2019.00683} {Dual student: Breaking the limits of the teacher in semi-supervised learning}.
\newblock In \emph{{ICCV}}, pages 6727--6735. {IEEE}.

\bibitem[{Kurakin et~al.(2017)Kurakin, Goodfellow, and Bengio}]{KurakinGB17a}
Alexey Kurakin, Ian~J. Goodfellow, and Samy Bengio. 2017.
\newblock \href {https://openreview.net/forum?id=HJGU3Rodl} {Adversarial examples in the physical world}.
\newblock In \emph{{ICLR}}. OpenReview.net.

\bibitem[{Laine and Aila(2017)}]{LaineA17}
Samuli Laine and Timo Aila. 2017.
\newblock \href {https://openreview.net/forum?id=BJ6oOfqge} {Temporal ensembling for semi-supervised learning}.
\newblock In \emph{{ICLR}}. OpenReview.net.

\bibitem[{Lee et~al.(2013)}]{lee2013pseudo}
Dong-Hyun Lee et~al. 2013.
\newblock \href {https://www.kaggle.com/blobs/download/forum-message-attachment-files/746/pseudo_label_final.pdf} {Pseudo-label: The simple and efficient semi-supervised learning method for deep neural networks}.
\newblock In \emph{ICML}, volume~3, page 896.

\bibitem[{Lee et~al.(2020)Lee, Hudson, Lee, and Manning}]{LeeHLM20}
Haejun Lee, Drew~A. Hudson, Kangwook Lee, and Christopher~D. Manning. 2020.
\newblock \href {https://doi.org/10.18653/v1/2020.emnlp-main.120} {{SLM:} learning a discourse language representation with sentence unshuffling}.
\newblock In \emph{{EMNLP}}, pages 1551--1562. Association for Computational Linguistics.

\bibitem[{Li et~al.(2022)Li, Fan, Song, Li, Li, Shao, and Zhan}]{LiFSLLYZ22}
Xin{-}Chun Li, Wen{-}Shu Fan, Shaoming Song, Yinchuan Li, Bingshuai Li, Yunfeng Shao, and De{-}Chuan Zhan. 2022.
\newblock Asymmetric temperature scaling makes larger networks teach well again.
\newblock In \emph{NeurIPS}.

\bibitem[{Lin and Hovy(2003)}]{LinH03}
Chin{-}Yew Lin and Eduard~H. Hovy. 2003.
\newblock \href {https://aclanthology.org/N03-1020/} {Automatic evaluation of summaries using n-gram co-occurrence statistics}.
\newblock In \emph{{HLT-NAACL}}.

\bibitem[{Liu et~al.(2021)Liu, Zhang, Fu, Hou, and Li}]{LiuZFHL21}
Chen Liu, Mengchao Zhang, Zhibing Fu, Panpan Hou, and Yu~Li. 2021.
\newblock \href {https://doi.org/10.18653/v1/2021.emnlp-main.192} {Flitext: {A} faster and lighter semi-supervised text classification with convolution networks}.
\newblock In \emph{{EMNLP}}, pages 2481--2491. Association for Computational Linguistics.

\bibitem[{Liu and Lapata(2019)}]{LiuL19}
Yang Liu and Mirella Lapata. 2019.
\newblock \href {https://doi.org/10.18653/v1/D19-1387} {Text summarization with pretrained encoders}.
\newblock In \emph{{EMNLP-IJCNLP}}, pages 3728--3738. Association for Computational Linguistics.

\bibitem[{Liu et~al.(2019)Liu, Ott, Goyal, Du, Joshi, Chen, Levy, Lewis, Zettlemoyer, and Stoyanov}]{abs190711692}
Yinhan Liu, Myle Ott, Naman Goyal, Jingfei Du, Mandar Joshi, Danqi Chen, Omer Levy, Mike Lewis, Luke Zettlemoyer, and Veselin Stoyanov. 2019.
\newblock \href {http://arxiv.org/abs/1907.11692} {Roberta: {A} robustly optimized {BERT} pretraining approach}.
\newblock \emph{CoRR}, abs/1907.11692.

\bibitem[{McAuley and Leskovec(2013)}]{McAuleyL13}
Julian~J. McAuley and Jure Leskovec. 2013.
\newblock \href {https://doi.org/10.1145/2507157.2507163} {Hidden factors and hidden topics: understanding rating dimensions with review text}.
\newblock In \emph{RecSys}, pages 165--172. {ACM}.

\bibitem[{Mendes et~al.(2012)Mendes, Jakob, and Bizer}]{mendes2012dbpedia}
Pablo~N Mendes, Max Jakob, and Christian Bizer. 2012.
\newblock \href {https://madoc.bib.uni-mannheim.de/59280/1/570_Paper.pdf} {\emph{DBpedia: A multilingual cross-domain knowledge base}}.
\newblock European Language Resources Association (ELRA).

\bibitem[{Mihalcea and Tarau(2004)}]{MihalceaT04}
Rada Mihalcea and Paul Tarau. 2004.
\newblock \href {https://aclanthology.org/W04-3252/} {Textrank: Bringing order into text}.
\newblock In \emph{{EMNLP}}, pages 404--411. Association for Computational Linguistics.

\bibitem[{Mirzadeh et~al.(2020)Mirzadeh, Farajtabar, Li, Levine, Matsukawa, and Ghasemzadeh}]{MirzadehFLLMG20}
Seyed{-}Iman Mirzadeh, Mehrdad Farajtabar, Ang Li, Nir Levine, Akihiro Matsukawa, and Hassan Ghasemzadeh. 2020.
\newblock \href {https://ojs.aaai.org/index.php/AAAI/article/view/5963} {Improved knowledge distillation via teacher assistant}.
\newblock In \emph{{AAAI}}, pages 5191--5198. {AAAI} Press.

\bibitem[{Miyato et~al.(2019)Miyato, Maeda, Koyama, and Ishii}]{MiyatoMKI19}
Takeru Miyato, Shin{-}ichi Maeda, Masanori Koyama, and Shin Ishii. 2019.
\newblock \href {https://doi.org/10.1109/TPAMI.2018.2858821} {Virtual adversarial training: {A} regularization method for supervised and semi-supervised learning}.
\newblock \emph{{IEEE} Trans. Pattern Anal. Mach. Intell.}, 41(8):1979--1993.

\bibitem[{Ouyang et~al.(2022)Ouyang, Wu, Jiang, Almeida, Wainwright, Mishkin, Zhang, Agarwal, Slama, Ray, Schulman, Hilton, Kelton, Miller, Simens, Askell, Welinder, Christiano, Leike, and Lowe}]{Ouyang0JAWMZASR22}
Long Ouyang, Jeffrey Wu, Xu~Jiang, Diogo Almeida, Carroll~L. Wainwright, Pamela Mishkin, Chong Zhang, Sandhini Agarwal, Katarina Slama, Alex Ray, John Schulman, Jacob Hilton, Fraser Kelton, Luke Miller, Maddie Simens, Amanda Askell, Peter Welinder, Paul~F. Christiano, Jan Leike, and Ryan Lowe. 2022.
\newblock \href {http://papers.nips.cc/paper\_files/paper/2022/hash/b1efde53be364a73914f58805a001731-Abstract-Conference.html} {Training language models to follow instructions with human feedback}.
\newblock In \emph{NeurIPS}.

\bibitem[{Park et~al.(2020)Park, Kim, You, and Cho}]{ParkKYC20}
Wonpyo Park, Wonjae Kim, Kihyun You, and Minsu Cho. 2020.
\newblock \href {https://doi.org/10.1007/978-3-030-66415-2\_49} {Diversified mutual learning for deep metric learning}.
\newblock In \emph{{ECCV}}, volume 12535 of \emph{Lecture Notes in Computer Science}, pages 709--725. Springer.

\bibitem[{Pereyra et~al.(2017)Pereyra, Tucker, Chorowski, Kaiser, and Hinton}]{PereyraTCKH17}
Gabriel Pereyra, George Tucker, Jan Chorowski, Lukasz Kaiser, and Geoffrey~E. Hinton. 2017.
\newblock \href {https://openreview.net/forum?id=HyhbYrGYe} {Regularizing neural networks by penalizing confident output distributions}.
\newblock In \emph{{ICLR}}. OpenReview.net.

\bibitem[{Qiao et~al.(2018)Qiao, Shen, Zhang, Wang, and Yuille}]{QiaoSZWY18}
Siyuan Qiao, Wei Shen, Zhishuai Zhang, Bo~Wang, and Alan~L. Yuille. 2018.
\newblock \href {https://doi.org/10.1007/978-3-030-01267-0\_9} {Deep co-training for semi-supervised image recognition}.
\newblock In \emph{{ECCV}}, volume 11219 of \emph{Lecture Notes in Computer Science}, pages 142--159. Springer.

\bibitem[{Radford et~al.(2018)Radford, Narasimhan, Salimans, Sutskever et~al.}]{radford2018improving}
Alec Radford, Karthik Narasimhan, Tim Salimans, Ilya Sutskever, et~al. 2018.
\newblock \href {https://www.cs.ubc.ca/~amuham01/LING530/papers/radford2018improving.pdf} {Improving language understanding by generative pre-training}.
\newblock \emph{OpenAI blog}.

\bibitem[{Radford et~al.(2019)Radford, Wu, Child, Luan, Amodei, Sutskever et~al.}]{radford2019language}
Alec Radford, Jeffrey Wu, Rewon Child, David Luan, Dario Amodei, Ilya Sutskever, et~al. 2019.
\newblock \href {https://openreview.net/forum?id=fylclEqgvgd} {Language models are unsupervised multitask learners}.
\newblock \emph{OpenAI blog}, 1(8):9.

\bibitem[{Rasmus et~al.(2015)Rasmus, Berglund, Honkala, Valpola, and Raiko}]{RasmusBHVR15}
Antti Rasmus, Mathias Berglund, Mikko Honkala, Harri Valpola, and Tapani Raiko. 2015.
\newblock \href {https://proceedings.neurips.cc/paper/2015/hash/378a063b8fdb1db941e34f4bde584c7d-Abstract.html} {Semi-supervised learning with ladder networks}.
\newblock In \emph{NeuralPS}, pages 3546--3554.

\bibitem[{Sajjadi et~al.(2016)Sajjadi, Javanmardi, and Tasdizen}]{SajjadiJT16}
Mehdi Sajjadi, Mehran Javanmardi, and Tolga Tasdizen. 2016.
\newblock \href {https://proceedings.neurips.cc/paper/2016/hash/30ef30b64204a3088a26bc2e6ecf7602-Abstract.html} {Regularization with stochastic transformations and perturbations for deep semi-supervised learning}.
\newblock In \emph{{NeurIPS}}, pages 1163--1171.

\bibitem[{Shen et~al.(2020)Shen, Zheng, Shen, Qu, and Chen}]{abs200913818}
Dinghan Shen, Mingzhi Zheng, Yelong Shen, Yanru Qu, and Weizhu Chen. 2020.
\newblock \href {https://arxiv.org/abs/2009.13818} {A simple but tough-to-beat data augmentation approach for natural language understanding and generation}.
\newblock \emph{CoRR}, abs/2009.13818.

\bibitem[{Shi et~al.(2023)Shi, Suzgun, Freitag, Wang, Srivats, Vosoughi, Chung, Tay, Ruder, Zhou, Das, and Wei}]{ShiSF0SVCTRZ0W23}
Freda Shi, Mirac Suzgun, Markus Freitag, Xuezhi Wang, Suraj Srivats, Soroush Vosoughi, Hyung~Won Chung, Yi~Tay, Sebastian Ruder, Denny Zhou, Dipanjan Das, and Jason Wei. 2023.
\newblock \href {https://openreview.net/pdf?id=fR3wGCk-IXp} {Language models are multilingual chain-of-thought reasoners}.
\newblock In \emph{{ICLR}}. OpenReview.net.

\bibitem[{Sohn et~al.(2020)Sohn, Berthelot, Carlini, Zhang, Zhang, Raffel, Cubuk, Kurakin, and Li}]{sohn2020fixmatch}
Kihyuk Sohn, David Berthelot, Nicholas Carlini, Zizhao Zhang, Han Zhang, Colin Raffel, Ekin~Dogus Cubuk, Alexey Kurakin, and Chun{-}Liang Li. 2020.
\newblock \href {https://proceedings.neurips.cc/paper/2020/hash/06964dce9addb1c5cb5d6e3d9838f733-Abstract.html} {Fixmatch: Simplifying semi-supervised learning with consistency and confidence}.
\newblock In \emph{NeurIPS}.

\bibitem[{Tarvainen and Valpola(2017)}]{TarvainenV17}
Antti Tarvainen and Harri Valpola. 2017.
\newblock \href {https://openreview.net/forum?id=ry8u21rtl} {Mean teachers are better role models: Weight-averaged consistency targets improve semi-supervised deep learning results}.
\newblock In \emph{{ICLR}}. OpenReview.net.

\bibitem[{Wang et~al.(2023{\natexlab{a}})Wang, Jiang, Zhong, and Liu}]{WangJZL23}
Chenyang Wang, Junjun Jiang, Zhiwei Zhong, and Xianming Liu. 2023{\natexlab{a}}.
\newblock \href {https://doi.org/10.1109/CVPR52729.2023.02141} {Spatial-frequency mutual learning for face super-resolution}.
\newblock In \emph{{CVPR}}, pages 22356--22366. {IEEE}.

\bibitem[{Wang et~al.(2023{\natexlab{b}})Wang, Zhu, and Wang}]{abs-2301-11916}
Xinyi Wang, Wanrong Zhu, and William~Yang Wang. 2023{\natexlab{b}}.
\newblock \href {https://doi.org/10.48550/ARXIV.2301.11916} {Large language models are implicitly topic models: Explaining and finding good demonstrations for in-context learning}.
\newblock \emph{CoRR}, abs/2301.11916.

\bibitem[{Wang et~al.(2022{\natexlab{a}})Wang, Chen, Fan, Sun, Tao, Hou, Wang, Yang, Zhou, Guo, Qi, Wu, Li, Nakamura, Ye, Savvides, Raj, Shinozaki, Schiele, Wang, Xie, and Zhang}]{WangCFSTHWY0GQW22}
Yidong Wang, Hao Chen, Yue Fan, Wang Sun, Ran Tao, Wenxin Hou, Renjie Wang, Linyi Yang, Zhi Zhou, Lan{-}Zhe Guo, Heli Qi, Zhen Wu, Yu{-}Feng Li, Satoshi Nakamura, Wei Ye, Marios Savvides, Bhiksha Raj, Takahiro Shinozaki, Bernt Schiele, Jindong Wang, Xing Xie, and Yue Zhang. 2022{\natexlab{a}}.
\newblock \href {http://papers.nips.cc/paper\_files/paper/2022/hash/190dd6a5735822f05646dc27decff19b-Abstract-Datasets\_and\_Benchmarks.html} {{USB:} {A} unified semi-supervised learning benchmark for classification}.
\newblock In \emph{NeurIPS}.

\bibitem[{Wang et~al.(2023{\natexlab{c}})Wang, Chen, Heng, Hou, Fan, Wu, Wang, Savvides, Shinozaki, Raj, Schiele, and Xie}]{Wang0HHFW0SSRS023}
Yidong Wang, Hao Chen, Qiang Heng, Wenxin Hou, Yue Fan, Zhen Wu, Jindong Wang, Marios Savvides, Takahiro Shinozaki, Bhiksha Raj, Bernt Schiele, and Xing Xie. 2023{\natexlab{c}}.
\newblock \href {https://openreview.net/pdf?id=PDrUPTXJI\_A} {Freematch: Self-adaptive thresholding for semi-supervised learning}.
\newblock In \emph{{ICLR}}. OpenReview.net.

\bibitem[{Wang et~al.(2022{\natexlab{b}})Wang, Mao, Liu, Jiang, Zhu, and Li}]{WangMLJZL22}
Yiming Wang, Qianren Mao, Junnan Liu, Weifeng Jiang, Hongdong Zhu, and Jianxin Li. 2022{\natexlab{b}}.
\newblock \href {https://aclanthology.org/2022.coling-1.561} {Noise-injected consistency training and entropy-constrained pseudo labeling for semi-supervised extractive summarization}.
\newblock In \emph{{COLING}}, pages 6447--6456. International Committee on Computational Linguistics.

\bibitem[{Wei et~al.(2022)Wei, Tay, Bommasani, Raffel, Zoph, Borgeaud, Yogatama, Bosma, Zhou, Metzler, Chi, Hashimoto, Vinyals, Liang, Dean, and Fedus}]{abs-2206-07682}
Jason Wei, Yi~Tay, Rishi Bommasani, Colin Raffel, Barret Zoph, Sebastian Borgeaud, Dani Yogatama, Maarten Bosma, Denny Zhou, Donald Metzler, Ed~H. Chi, Tatsunori Hashimoto, Oriol Vinyals, Percy Liang, Jeff Dean, and William Fedus. 2022.
\newblock \href {https://doi.org/10.48550/arXiv.2206.07682} {Emergent abilities of large language models}.
\newblock \emph{CoRR}, abs/2206.07682.

\bibitem[{Xie et~al.(2020)Xie, Dai, Hovy, Luong, and Le}]{XieDHL020}
Qizhe Xie, Zihang Dai, Eduard~H. Hovy, Thang Luong, and Quoc Le. 2020.
\newblock \href {https://proceedings.neurips.cc/paper/2020/hash/44feb0096faa8326192570788b38c1d1-Abstract.html} {Unsupervised data augmentation for consistency training}.
\newblock In \emph{{NeurIPS}}.

\bibitem[{Xu et~al.(2022)Xu, Liu, and Abbasnejad}]{XuLA22}
Hai{-}Ming Xu, Lingqiao Liu, and Ehsan Abbasnejad. 2022.
\newblock \href {https://doi.org/10.18653/v1/2022.naacl-main.219} {Progressive class semantic matching for semi-supervised text classification}.
\newblock In \emph{{NAACL}}, pages 3003--3013. Association for Computational Linguistics.

\bibitem[{Yan et~al.(2021)Yan, Li, Wang, Zhang, Wu, and Xu}]{YanLWZWX20}
Yuanmeng Yan, Rumei Li, Sirui Wang, Fuzheng Zhang, Wei Wu, and Weiran Xu. 2021.
\newblock \href {https://doi.org/10.18653/v1/2021.acl-long.393} {Consert: {A} contrastive framework for self-supervised sentence representation transfer}.
\newblock In \emph{{ACL/IJCNLP}}, pages 5065--5075. Association for Computational Linguistics.

\bibitem[{Zhang et~al.(2021)Zhang, Wang, Hou, Wu, Wang, Okumura, and Shinozaki}]{ZhangWHWWOS21}
Bowen Zhang, Yidong Wang, Wenxin Hou, Hao Wu, Jindong Wang, Manabu Okumura, and Takahiro Shinozaki. 2021.
\newblock \href {https://proceedings.neurips.cc/paper/2021/hash/995693c15f439e3d189b06e89d145dd5-Abstract.html} {Flexmatch: Boosting semi-supervised learning with curriculum pseudo labeling}.
\newblock In \emph{NeurIPS}, pages 18408--18419.

\bibitem[{Zhang et~al.(2022{\natexlab{a}})Zhang, Wang, Campos, Huang, Guo, and Yang}]{ZhangW0HG022}
Miao Zhang, Li~Wang, David Campos, Wei Huang, Chenjuan Guo, and Bin Yang. 2022{\natexlab{a}}.
\newblock \href {http://papers.nips.cc/paper\_files/paper/2022/hash/4b25c000967af9036fb9b207b198a626-Abstract-Conference.html} {Weighted mutual learning with diversity-driven model compression}.
\newblock In \emph{NeurIPS}.

\bibitem[{Zhang et~al.(2022{\natexlab{b}})Zhang, Niranjan, and He}]{ZhangNH22}
Minjia Zhang, Uma{-}Naresh Niranjan, and Yuxiong He. 2022{\natexlab{b}}.
\newblock \href {https://doi.org/10.1609/AAAI.V36I10.21423} {Adversarial data augmentation for task-specific knowledge distillation of pre-trained transformers}.
\newblock In \emph{{AAAI}}, pages 11685--11693. {AAAI} Press.

\bibitem[{Zhang et~al.(2015)Zhang, Zhao, and LeCun}]{ZhangZL15}
Xiang Zhang, Junbo~Jake Zhao, and Yann LeCun. 2015.
\newblock \href {https://proceedings.neurips.cc/paper/2015/hash/250cf8b51c773f3f8dc8b4be867a9a02-Abstract.html} {Character-level convolutional networks for text classification}.
\newblock In \emph{{NeurIPS}}, pages 649--657.

\bibitem[{Zhang et~al.(2018)Zhang, Xiang, Hospedales, and Lu}]{ZhangXHL18}
Ying Zhang, Tao Xiang, Timothy~M. Hospedales, and Huchuan Lu. 2018.
\newblock \href {http://openaccess.thecvf.com/content\_cvpr\_2018/html/Zhang\_Deep\_Mutual\_Learning\_CVPR\_2018\_paper.html} {Deep mutual learning}.
\newblock In \emph{{CVPR}}, pages 4320--4328. Computer Vision Foundation / {IEEE} Computer Society.

\bibitem[{Zheng et~al.(2022)Zheng, You, Huang, Wang, Qian, and Xu}]{ZhengYHWQX22}
Mingkai Zheng, Shan You, Lang Huang, Fei Wang, Chen Qian, and Chang Xu. 2022.
\newblock \href {https://doi.org/10.1109/CVPR52688.2022.01407} {Simmatch: Semi-supervised learning with similarity matching}.
\newblock In \emph{{CVPR}}, pages 14451--14461. {IEEE}.

\bibitem[{Zheng et~al.(2024)Zheng, Zhang, Zhang, Ye, Luo, and Ma}]{abs-2403-13372}
Yaowei Zheng, Richong Zhang, Junhao Zhang, Yanhan Ye, Zheyan Luo, and Yongqiang Ma. 2024.
\newblock Llamafactory: Unified efficient fine-tuning of 100+ language models.
\newblock \emph{CoRR}, abs/2403.13372.

\bibitem[{Zhu et~al.(2023)Zhu, Li, Liu, Ma, and Wang}]{abs-2308-07633}
Xunyu Zhu, Jian Li, Yong Liu, Can Ma, and Weiping Wang. 2023.
\newblock \href {https://doi.org/10.48550/ARXIV.2308.07633} {A survey on model compression for large language models}.
\newblock \emph{CoRR}, abs/2308.07633.

\bibitem[{Zhu and Wang(2021)}]{ZhuW21a}
Yichen Zhu and Yi~Wang. 2021.
\newblock \href {https://doi.org/10.1109/ICCV48922.2021.00501} {Student customized knowledge distillation: Bridging the gap between student and teacher}.
\newblock In \emph{{ICCV}}, pages 5037--5046. {IEEE}.

\end{thebibliography}

\newpage
\appendix

\section{Appendix}
\label{sec:appendix}

\subsection{Baselines Details}
\label{Baselines_Details}
For the text classification task, TinyBERT~\cite{JiaoYSJCL0L20} is a compressed model implemented by 6-layer or 4-layer BERT\(\rm_{BASE}\). 
For semi-supervised methods, we use the released code to train the UDA, which includes ready-made 12-layer BERT\(\rm_{BASE}\), 6-layer, or 4-layer TinyBERT. 
In extractive summarization, the ORACLE system serves as an upper bound in the domain of extractive summarization.

Other SSL algorithms integrated with BERT are implemented in a unified semi-supervised learning benchmark (USB)~\cite{WangCFSTHWY0GQW22}, including  Mean Teacher~\cite{TarvainenV17}, VAT~\cite{MiyatoMKI19}, FixMatch~\cite{sohn2020fixmatch}, AdaMatch~\cite{BerthelotRSCK22}, and SimMatch~\cite{ZhengYHWQX22}. 
These methods boost model robustness by ensuring consistent predictions on perturbed unlabeled samples, or uses pseudo labels for training enhancement.
PCM~\cite{XuLA22} is a complex multi-submodule combination SSL model with a 12-layer BERT backbone.
Recently,  FreeMatch~\cite{Wang0HHFW0SSRS023} dynamically adjusts the confidence threshold based on the model's learning status. SoftMatch~\cite{abs-2301-10921} employs a unified sample weighting formulation for pseudo-labeling. The majority of models were originally introduced in the realm of computer vision, and we present their text classification outcomes in the USB benchmark evaluation. Fully-Svp merges unsupervised data labels with labeled data for BERT training, while Svp is solely on BERT's classification results from labeled data.


\subsection{Model Efficiency Analysis}
\label{Model_Efficiency}
\Disco and our PS-NET employ student models, such as 4-layer or 2-layer BERT. In other words, during the model deployment phase, \Disco and PS-NET demonstrate identical inference speeds of 4-layer or 2-layer BERT. 
As shown in Table~\ref{model_efficiency}, 
compared with the teacher BERT\(_{\rm BASE}\),  the 2-layer small models are 12.30\(\times\) smaller and 7.52\(\times\) inference speedup in the model efficiency.  
FLiText is slightly faster than the smaller model generated \Disco and PS-NET. This is because FLiText uses a convolutional network while our student models use BERT with multi-head self-attention. The lower computational complexity of convolutional networks. The 1D-CNN requires \(O(k\times n\times d)\) operations\footnote{\(n\) is the sequence length, \(d\) is the representation dimension, \(k\) is the kernel size of convolutions.} used by FLiText. In contrast, the multi-head self-attention mechanism of BERT requires \(O(n^{2}\times d + n\times d^{2})\) operations. However, despite the FLiText-model having more parameters, it gives a worse performance compared to the smaller student model generated by \Disco and PS-NET in Table~\ref{main_text_classification}. Our PS-NET achieves optimal performance and maintains comparable inference speed.

\begin{table}[t]
  \centering
  \footnotesize
  \caption{Model efficiency about inference speedup on a single NVIDIA Tesla V100 32GB GPU.  \textbf{\(\mathcal{T}_{\rm TS}\)}(ms) refers to the speedup of extractive summarization models trained with 100 labelled data. \textbf{\(\mathcal{T}_{\rm TC}\)}(ms) illustrates the speedup of text classification models trained with AG News 200 labelled data per class.}
    \renewcommand\arraystretch{1.2}
    \setlength{\abovecaptionskip}{0.5mm}
    \setlength{\tabcolsep}{1.1mm}{
    \begin{tabular}{@{}lr|lr@{}}
    \toprule
    \textbf{Models}
    & \multicolumn{1}{c}{\textbf{\(\mathcal{T}_{\rm TS}\)}(ms)}
    & \multicolumn{1}{|l}{\textbf{Models}}
    & \multicolumn{1}{r@{}}{\textbf{\(\mathcal{T}_{\rm TC}\)}(ms)}  \\
    \midrule
    BERTSUM
    & 12.66
    &  BERT\(\rm_{BASE}\)
    & 12.94 \\
    CPSUM
     & 12.66
     & TinyBERT\(^{4}\)
    & 2.86 \\
    TinyBERTSUM\(^{4}\)
     & 2.64
     &  UDA\(\rm_{TinyBERT^{4}}\)
    & 2.86 \\
    UDASUM\(\rm_{TinyBERT^{4}}\)
     & 2.64
     & FLiText
    & 1.56 \\
    \Disco {\scriptsize  ($\rm S^{A2}$ or $\rm S^{B2}$)}
     & 1.72
     & \Disco {\scriptsize  ($\rm S^{A2}$ or $\rm S^{B2}$)}
    & 1.72 \\
    \midrule
    PS-NET {\scriptsize ($\rm S^{A2}$ or $\rm S^{B2}$)}
     & 1.72
     & PS-NET {\scriptsize ($\rm S^{A2}$ or $\rm S^{B2}$)}
    & 1.72 \\
    \bottomrule
    \end{tabular}%
    }
  \label{model_efficiency}%
\end{table}%

\begin{figure*}[htbp]
  \centering
  \begin{minipage}[t]{0.24\linewidth}
  \centering
  \subfigure[STR-1 vs STR-2]{
    \includegraphics[width=1.45in]{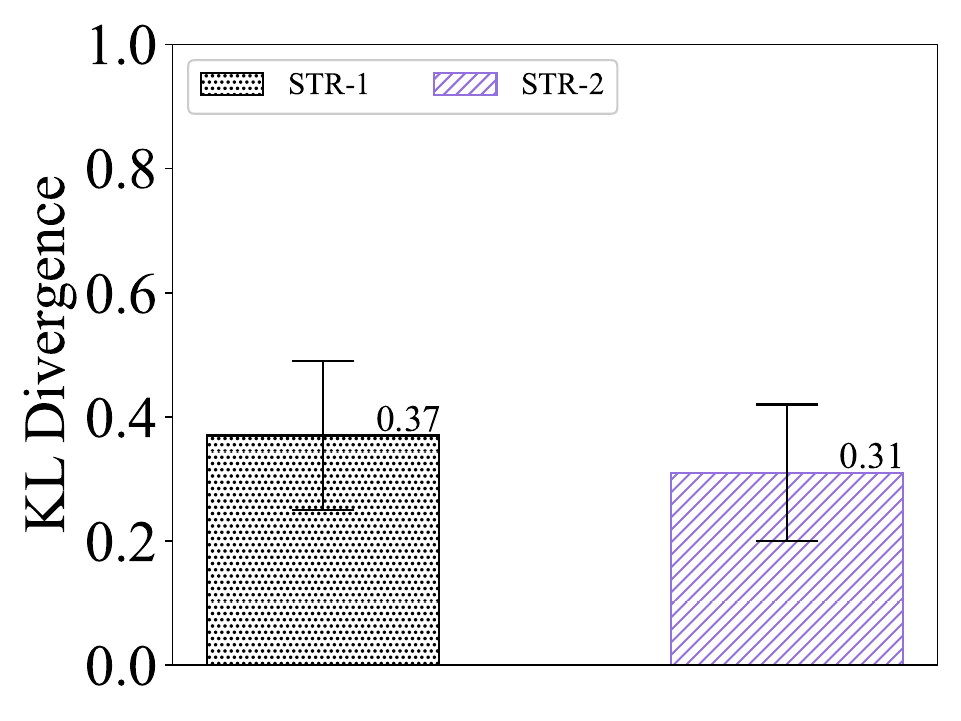}}
  \end{minipage}%
  \begin{minipage}[t]{0.24\linewidth}
  \centering
      \subfigure[STR-1 vs STR-3]{
  \includegraphics[width=1.45in]{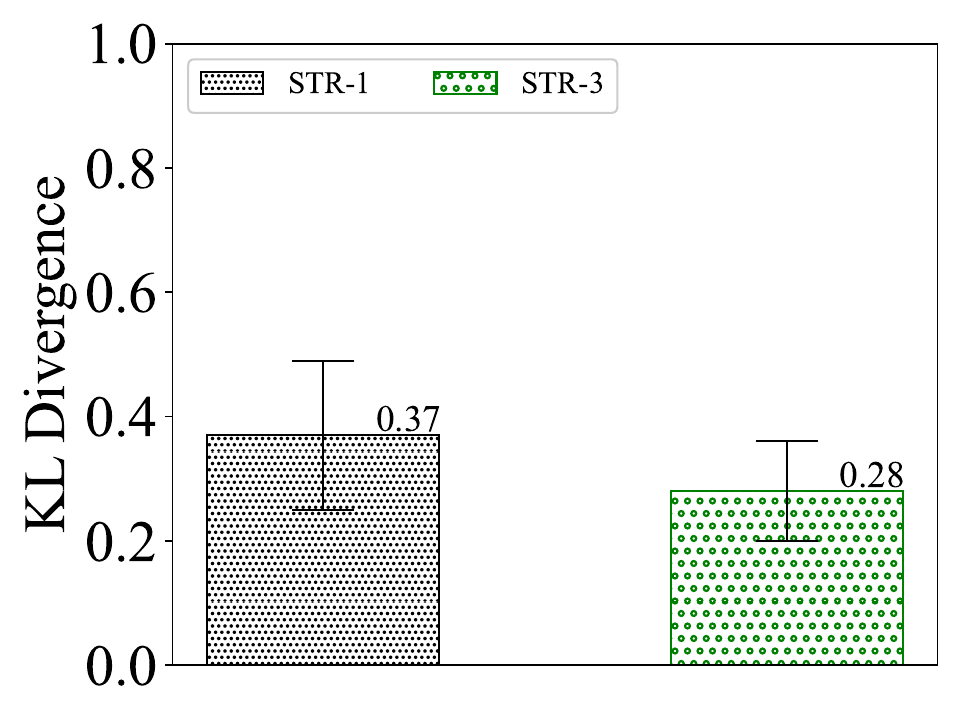}}
      \label{EGAT}
  \end{minipage}
    \begin{minipage}[t]{0.24\linewidth}
  \centering
      \subfigure[STR-1 vs STR-4.1]{
  \includegraphics[width=1.45in]{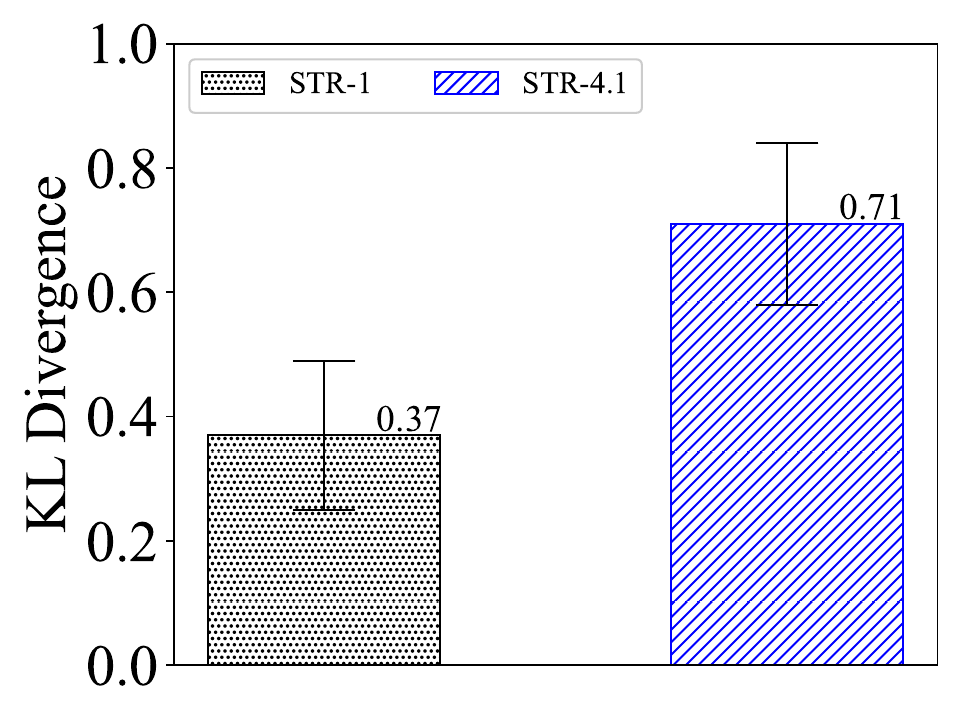}}
      \label{NGAT}
  \end{minipage}
    \begin{minipage}[t]{0.24\linewidth}
  \centering
      \subfigure[STR-1 vs STR-4.2]{
  \includegraphics[width=1.45in]{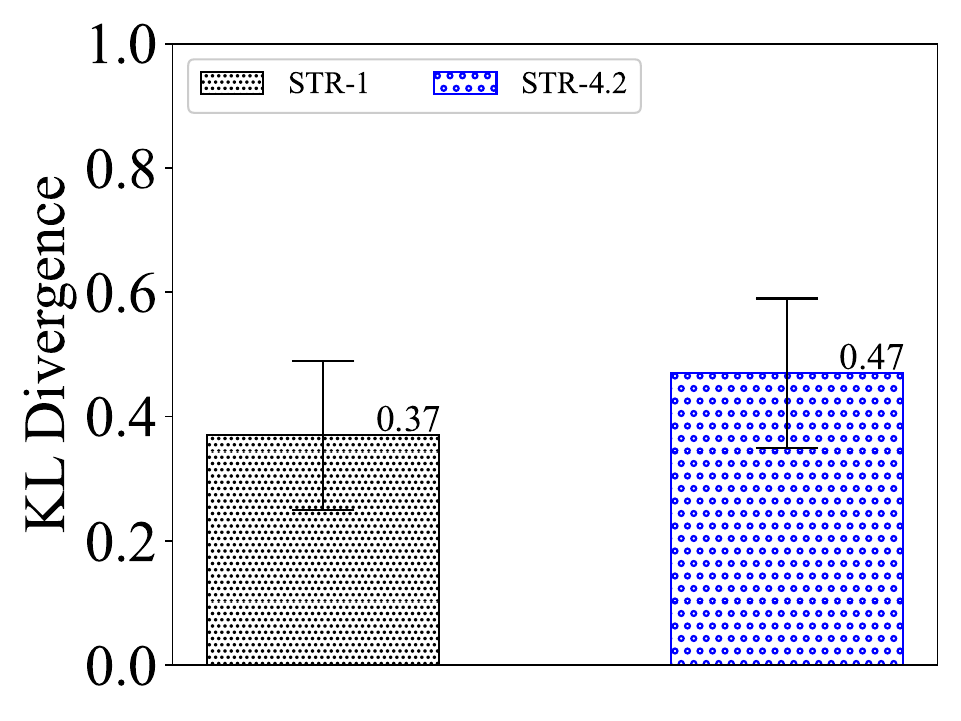}}
      \label{ENGAT}
  \end{minipage}
      \vspace{-0.1in}
     \caption{Match manner of KL divergence on PS-NET teacher and students. A smaller KL divergence value indicates less mismatch. Strategy 1 (STR-1) employs two 2-layer students in PS-NET, utilizing the first 2 layers and the last 2 layers of BERT teacher, respectively. Strategy 2 (STR-2) involves two 6-layer students in PS-NET. Strategy 3 (STR-3) utilizes six 2-layer students in PS-NET. Strategy 4.1 (STR-4.1) and Strategy 4.2 (STR-4.2) incorporate two SingleStudent models and BERT teacher, with two students corresponding to the first 2 layers (STR-4.1) and the last 2 layers (STR-4.2) of BERT teacher. }
    \label{loss_2D}
\end{figure*}

\begin{table}[tb]
  \footnotesize
 \caption{Validation accuracy (Acc (\%)) comparison between PS-NET, \Disco students and DML students (training from scratch) in SSL text classifications only using a limited 10 labelled data per class.}
        \renewcommand\arraystretch{1.2}
     \setlength{\abovecaptionskip}{0.5mm}
     \setlength{\tabcolsep}{2.2mm}{
\begin{tabular}{@{}l|c|c|r@{}}
\toprule
\textbf{Models}                                            & \textbf{AG News}        & \textbf{Yahoo!Answer}   & \textbf{DBpedia}        \\ \hline
DML(Net A2) & 34.92 &17.24 &54.86 \\
DML(Net B2) & 34.82 &17.46 &61.28 \\
DML(Net A4) & 37.51 &21.23 &65.28 \\
DML(Net B4) & 36.76 &21.62 &64.80 \\
\Disco (S\(^{\rm A2}\))  & 70.47          & 48.10          & 89.78          \\
\Disco (S\(^{\rm B2}\))   & 74.30          & 50.95          & 89.80          \\
PS-NET (S\(^{\rm A2}\))   & 81.14          & 61.12          & 96.61          \\
PS-NET (S\(^{\rm B2}\))   & \textbf{81.89} & \textbf{63.91} & \textbf{98.05} \\ \cline{1-2} \cline{1-4} 
\Disco (S\(^{\rm A6}\))  & 74.69          & 57.42          & 98.06          \\
\Disco (S\(^{\rm B6}\))   & 77.40          & 58.48          & 98.03 \\
PS-NET (S\(^{\rm A6}\))    & 82.39          & 64.12          & 98.51          \\
PS-NET (S\(^{\rm B6}\))  & \textbf{82.61} & \textbf{64.33} & \textbf{98.49}          \\ \bottomrule
\end{tabular}%
     }
     \label{muse_disco}
\end{table}

\begin{table}[t]
  \centering
  \footnotesize
  \caption{Test accuracy (Acc (\%)) of other prominent SSL  models and our PS-NET. All results are reported by the Unified SSL Benchmark (USB)~\cite{WangCFSTHWY0GQW22}. L\(_{m}\) is the number of the BERT layers.}
     \renewcommand\arraystretch{1.2}
  \setlength{\abovecaptionskip}{2mm}
  \setlength{\tabcolsep}{1.1mm}{
    \begin{tabular}{@{}l|l|l|r|l}
    \toprule
    \(\mathcal{D}\) & \textbf{Models} & \textbf{L}\(_{m}\) & \textbf{L}\(_{d}\) & Acc \\
    \hline
    \multirow{6}[3]{*}{\rotatebox{90}{\textbf{AG News}}} & \(\prod \)-model{\scriptsize~\cite{RasmusBHVR15}}  & \multicolumn{1}{c|}{\multirow{6}[3]{*}{12}} & \cellcolor{gray!5}50   &   \cellcolor{gray!5}86.56 
    \\        
& P-Labeling{\scriptsize~\cite{lee2013pseudo}}   &       & \cellcolor{gray!5}50    & \cellcolor{gray!5}87.01
\\        
& MeanTeacher{\scriptsize~\cite{TarvainenV17}}   &       & \cellcolor{gray!5}50    & \cellcolor{gray!5}86.77 
\\        
& PCM{\scriptsize~\cite{XuLA22}}  &       & \cellcolor{gray!5}50    & \cellcolor{gray!5}\textbf{88.85} \\
          & MixText{\scriptsize~\cite{ChenYY20}} &       & \cellcolor{gray!15}30    & \cellcolor{gray!15}87.40 \\
\cline{2-5}          
& \Disco  & \multicolumn{1}{c|}{6} & \cellcolor{gray!15}30    &  \cellcolor{gray!15}86.93 \\ 
\cline{2-5}          
& PS-NET (ours) & \multicolumn{1}{c|}{2} & \cellcolor{gray!15}30    &  \cellcolor{gray!15}\textbf{87.53} \\  
\cline{1-5}
    \multirow{6}[26]{*}{\rotatebox{90}{\textbf{Yahoo!Answer}}} 
    & P-Labeling{\scriptsize~\cite{lee2013pseudo}}   &  \multicolumn{1}{c|}{\multirow{6}[24]{*}{12}}     & \cellcolor{gray!40}200    & \cellcolor{gray!40}66.56
\\        
& MeanTeacher{\scriptsize~\cite{TarvainenV17}}   &       & \cellcolor{gray!40}200    & \cellcolor{gray!40}66.57 
\\  
& \(\prod \)-model{\scriptsize~\cite{RasmusBHVR15}} &  & \cellcolor{gray!40}200  &  \cellcolor{gray!40}67.04 \\      
& VAT{\scriptsize~\cite{MiyatoMKI19}}   &       & \cellcolor{gray!40}200  &  \cellcolor{gray!40}68.47 \\
        & FlexMatch{\scriptsize~\cite{ZhangWHWWOS21}} &       & \cellcolor{gray!40}200  &  \cellcolor{gray!40}68.58 \\ 
    & AdaMatch{\scriptsize~\cite{BerthelotRSCK22}} &  & \cellcolor{gray!40}200  &  \cellcolor{gray!40}69.18 \\
    & FixMatch{\scriptsize~\cite{sohn2020fixmatch}} &       & \cellcolor{gray!40}200  &  \cellcolor{gray!40}69.24 \\
    & SimMatch{\scriptsize~\cite{ZhengYHWQX22}} &       & \cellcolor{gray!40}200  &  \cellcolor{gray!40}69.36 \\
    & CRMactch{\scriptsize~\cite{DBLP:journals/ijcv/FanKDS23/revisitconsistency}} &       & \cellcolor{gray!40}200  &  \cellcolor{gray!40}69.38 \\ 
    & SoftMactch{\scriptsize~\cite{abs-2301-10921}} &       & \cellcolor{gray!40}200  &  \cellcolor{gray!40}69.56 \\ 
    &  FreeMactch{\scriptsize~\cite{Wang0HHFW0SSRS023}} &       & \cellcolor{gray!40}200  &  \cellcolor{gray!40}69.68 \\  
               & SoftMatch{\scriptsize~\cite{abs-2301-10921}} &       & \cellcolor{gray!40}200  &  \cellcolor{gray!40}69.56 \\
\cline{2-5}          
& \Disco & \multicolumn{1}{c|}{6} & \cellcolor{gray!40}200   &  \cellcolor{gray!40}69.75 \\ 
\cline{2-5}
& PS-NET (ours) & \multicolumn{1}{c|}{4} & \cellcolor{gray!40}200    &  \cellcolor{gray!40}\textbf{71.24} \\  
    \bottomrule
    \end{tabular}%
    }
  \label{auxiliary_text_classification_2}
\end{table}%

\subsection{Match Manner Analysis}
We computed the match manner between the teacher network and student network using the average KL divergence (\(\tau\) = 1) on the predicted probabilities. Figure~\ref{loss_2D} (a) demonstrates that smaller representational capacity hinders student performance. A comparison between Figure~\ref{loss_2D} (a) and Figure~\ref{loss_2D} (b) reveals that increasing the number of students further diminishes the gap with the teacher network. In the context of a single student and a single teacher, shallow networks (such as STR-1) may amplify the discrepancy of predictions to the teacher and tend to be fairly severe. 

\subsection{Performance Superiority of PS-NET}
Regarding the validation experiment with a limited 10 labelled data per class, as depicted in Table~\ref{muse_disco}. PS-NET demonstrates superior performance compared to \Disco using the same training data. Besides, to further clarify the differences between our method and DML training from scratch, we compared the effectiveness of PS-NET  and pure DML under an extreme SSL setting (4-layer and 2-layer BERT, 10 labelled data per class). It can be seen that in Table~\ref{muse_disco}, with 10 labelled data per class, DML's performance barely exceeds that of a single model with random initialization and lags significantly behind PS-NET and other SSL baselines. The experimental results indicate that DML performs poorly when directly applied to SSL scenarios.   In contrast, our method is specifically designed for SSL, emphasizing that a more powerful teacher network is essential to guide student models in scenarios with sparsely labelled data.


Besides, Table~\ref{auxiliary_text_classification_2} provides more baseline SSL methods from the Unified SSL Benchmark (USB)~\cite{WangCFSTHWY0GQW22} in AG News, Yahoo!Answer datasets.  With 200 labelled examples per class, PS-NET with 4 BERT layers achieves 71.24\% accuracy, surpassing methods like VAT, FlexMatch, AdaMatch, and FixMatch, which use 12 BERT layers. With only 2 BERT layers and a relatively low labelled data per class (30), PS-NET also outperforms models like 6 BERT layers of \Disco. All supplementary experiments confirm the superior performance of the lightweight models achieved by PS-NET.



Finally, we sum up PS-NET integrates knowledge distillation and knowledge optimization within a unified framework, facilitating direct optimization of transferred knowledge by supervised signals.  
Moreover, in PS-NET, DML with interaction behaviour and CAT  with adversarial perturbations enable each network to acquire distinct knowledge, bolstering the generalization and robustness of lightweight models.

\subsection{Framework Variations from \Disco}
Both PS-NET and \Disco represent technical solutions for faster and lighter semi-supervised learning. Additionally, similarities with \Disco are apparent in the use of a semi-supervised learning framework, with multiple networks learning logits estimates collaboratively. Regarding semi-supervised optimization, the objectives of both models remain consistent. They aim to augment generalization ability by employing complementary learning from multiple peer networks to elevate the posterior entropy~\cite{PereyraTCKH17} of each student network, thereby fostering shared learning experiences.

The semi-supervised learning framework in \Disco originates from Deep Co-Training, DCO~\cite{QiaoSZWY18} which emphasizes multi-view learning. However, PS-NET's framework is derived from Deep Mutual Learning, DML~\cite{ZhangXHL18} which focuses on knowledge distillation variants. 
More information is provided in the original paper by \citet{QiaoSZWY18} and ~\citet{ZhangXHL18}. The proposed PS-NET differs from \Disco in the following aspects:
\begin{itemize}[leftmargin=*]
  \item \textbf{Distillation Methods}: \Disco uses offline distillation, necessitating pre-training of potent teacher networks and distillation of multiple students in advance. In contrast, PS-NET employs online distillation, where supervised learning and unsupervised distillation for both the teacher and student occur simultaneously. This design facilitates a more seamless emulation of the teacher's behavior. Specifically, the general knowledge is acquired during the initial training phases, while task-specific knowledge is learned in subsequent stages. This consistent emulation allows the teacher to guide the optimization paths of all students, rather than relying solely on peer learning without external guidance. By emphasizing external guidance, our framework enhances mutual learning between two divergent students, ultimately strengthening their collaborative learning process. 
  \item \textbf{Learning Procedures}: \Disco pre-distils multiple students and then conducts co-training, where labelled and unlabeled data are input together. In contrast, PS-NET performs supervised learning \(\textcircled{1}\) sequentially followed by unsupervised knowledge distillation. This approach encourages the teacher model to actively participate in every single optimization step, thereby mitigating the impact of the scale gap between the teacher and student models on distillation performance. Similar discussions can be found in methodologies such as TAKD~\cite{MirzadehFLLMG20} and BANs~\cite{FurlanelloLTIA18}. Furthermore, PS-NET incorporates  curriculum adversarial training (CAT) (shown in Algorithm 1) to \(\textcircled{2}\) progressively increase learning complexity. These procedures enables PS-NET to implement an iterative learning approach, facilitating continuous self-improvement of the lightweight model.
\end{itemize}



\end{document}